\documentclass[10pt,twocolumn,letterpaper]{article}

\usepackage{iccv}
\usepackage{times}
\usepackage{epsfig}
\usepackage{graphicx}
\usepackage{amsmath}
\usepackage{amssymb}
\usepackage{appendix}
\usepackage{pifont}
\usepackage{multirow}
\usepackage{bigstrut}
\usepackage{graphicx}
\usepackage{amsmath}
\usepackage{amssymb}
\usepackage{bbm}
\usepackage{booktabs}
\usepackage{algpseudocode}
\usepackage{algorithmicx,algorithm}
\usepackage{times}
\usepackage{epsfig}
\usepackage{appendix}
\usepackage{mathptm}
\usepackage{multirow}
\usepackage{booktabs} 
\usepackage[table,xcdraw]{xcolor}
\usepackage{color}
\newcommand{\red}[1]{\textcolor[RGB]{255,0,0}{#1}}

\usepackage{esint}
\usepackage{newtxmath}
\usepackage{textcomp}
\usepackage{caption}
\usepackage[pagebackref=true,breaklinks=true,letterpaper=true,colorlinks,bookmarks=false]{hyperref}

\iccvfinalcopy % *** Uncomment this line for the final submission

\def\iccvPaperID{1290} % *** Enter the ICCV Paper ID here

\ificcvfinal\fi

\begin{document}

%%%%%%%%% TITLE
\title{Detecting the open-world objects with the help of the ``Brain''}
% \title{The Metric for Unknown Objects Without Grounding Truth in Open World }
\author{
    Shuailei Ma \textsuperscript{\rm 1} ~
    Yuefeng Wang\textsuperscript{\rm 1} ~
    Ying Wei\textsuperscript{\rm 1 2} ~ 
    Peihao Chen \textsuperscript{\rm 3} \\
    Zhixiang Ye \textsuperscript{\rm 4} ~
    Jiaqi Fan\textsuperscript{\rm 1}~
    Enming Zhang\textsuperscript{\rm 1} ~
    Thomas H. Li\textsuperscript{\rm 5} ~
     \\
    \textsuperscript{\rm 1}\small{Northeast University, Shenyang, China}
    \textsuperscript{\rm 2}\small{Information Technology R\&D Innovation Center of Peking University} \\
    \textsuperscript{\rm 3}\small{South China University of Technology,}
    \textsuperscript{\rm 4}\small{Nanjing University of Information Science and Technology} \\
    \textsuperscript{\rm 5}\small{School of Electronic and Computer Engineering, Peking University Shenzhen Graduate School, Shenzhen, China}\\
    % mingkuitan@scut.edu.cn, tli@aiit.org.cn \\
    % \{csxinyusun, phchencs, lwchencs, ganchuang1990\}@gmail.com
}
\twocolumn[{%
\renewcommand\twocolumn[1][]{#1}%
\maketitle
\begin{center}
\includegraphics[width=\textwidth]{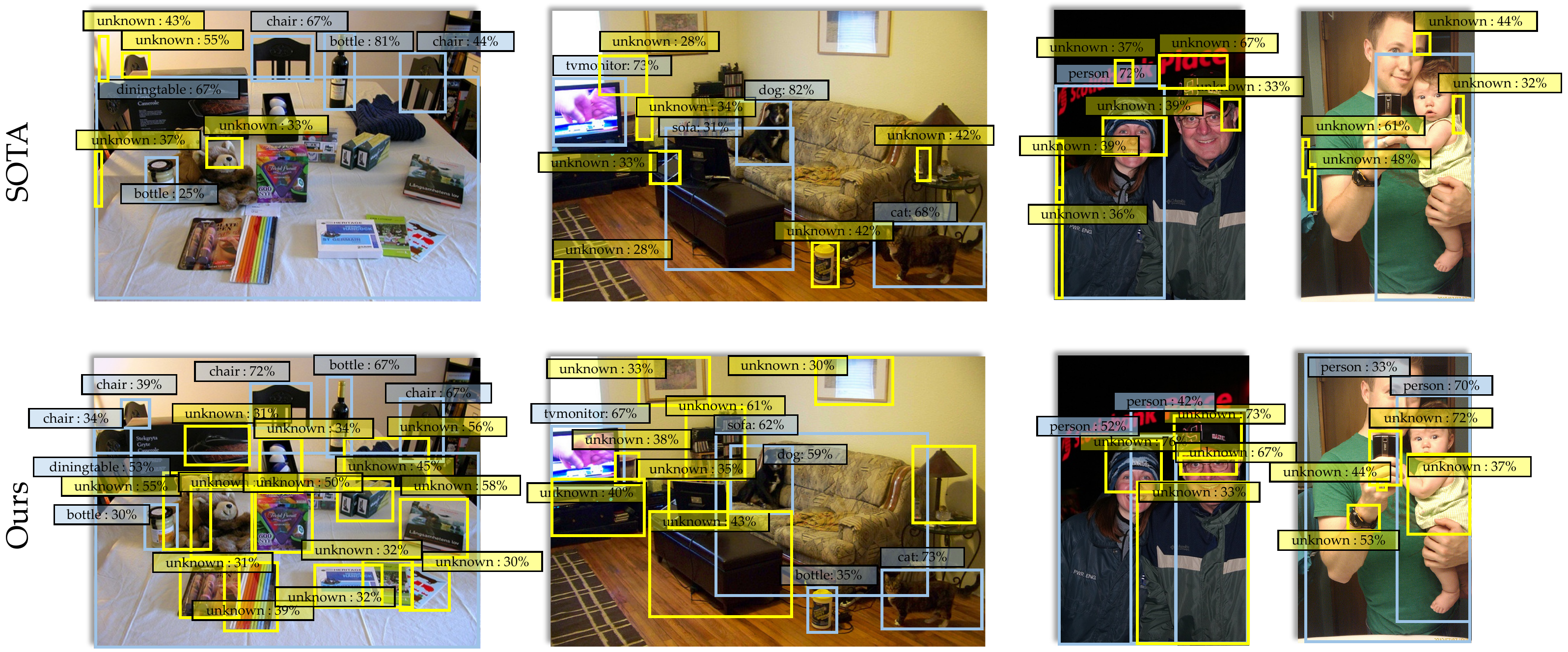}
\captionof{figure}{Visualization results for the comparison between the SOTA method and our model (known objects in \textcolor[rgb]{0.1,0.8,0.9}{Blue}
 and unknown objects in \textcolor[RGB]{255,215,0}{Yellow}). The categories of PASCAL VOC \cite{voc} is set as known.  Our model significantly outperforms the SOTA (OW-DETR \cite{owdetr}) for open-world object detection, almost accurately detecting all unknown objects in the open scene, including \textit{clothes} and \textit{hats} worn by people, while most of the unknown objects detected by the SOTA model are backgrounds.} 
\end{center}%
}]
\maketitle
\ificcvfinal\thispagestyle{empty}\fi

%%%%%%%%% ABSTRACT
\begin{abstract}
% \vspace{-0.3cm}
Open World Object Detection (OWOD) is a novel computer vision task with a considerable challenge, bridging the gap between classic object detection (OD) benchmarks and real-world object detection. In addition to detecting and classifying seen/known objects, OWOD algorithms are expected to detect unseen/unknown objects and incrementally learn them. The natural instinct of humans to identify unknown objects in their environments mainly depends on their brains' knowledge base. It is difficult for a model to do this only by learning from the annotation of several tiny datasets. The large pre-trained grounded language-image models - VL (\ie GLIP) have rich knowledge about the open world but are limited to the text prompt. We propose leveraging the VL as the ``Brain'' of the open-world detector by simply generating unknown labels. Leveraging it is non-trivial because the unknown labels impair the model's learning of known objects. In this paper, we alleviate these problems by proposing the \textbf{down-weight loss function} and \textbf{decoupled detection structure}. Moreover, our detector leverages the ``Brain'' to learn novel objects beyond VL through our \textbf{pseudo-labeling scheme}. Comprehensive experiments performed on MS-COCO and PASCAL VOC (\textbf{Ours results in the bolded font}) show that our model significantly outperforms the existing open-world detectors in terms of U-Recall on two existing splits ( 12.1, 9.4, 11.6 \vs \textbf{39.0, 36.7, 36.1} in ORE split, 5.7, 6.2, 6.9 \vs \textbf{60.9, 60.0, 58.6} in OW-DETR split ), without loss of performance on known classes. Compared with the GLIP, our model still achieves better performance ( 37.0, 35.5, 34.9 \vs \textbf{39.0, 36.7, 36.1} in ORE split, 52.6, 54.5, 53.3 \vs \textbf{60.9, 60.0, 58.6} in OW-DETR split ) for open-world object detection, with a faster inference speed ( 9.22 seconds / per image \vs \textbf{0.08 seconds / per image} ) and less computational cost ( FLOPs 965GMac \vs \textbf{212GMac}, Param\# 321.9M \vs \textbf{42.9M}). The code is available at \url{https://github.com/xiaomabufei/DOWB}.

% Despite the low speed and huge costs, they are limited to text prompts, which cannot include all open-world categories. In this paper, we leverage the zero-shot as the "Brain" to assist the model's learning. 
% To alleviate the problems posed by the co-existence of knowledge of the brain and data annotations, we utilize the label confidence to down-weight the training loss of the assistant knowledge. To equip the model to evolve unknown classes beyond the "Brain", we leverage a novel box score branch to select pseudo labels that make models exploratory. Compared with the assistant Zero-Shot detector, our model exceeds it in all tasks 
% the state-of-the-art methods. Remarkably, using minimal training resources, our model even exceeds the zero-shot detector for open-world object detection. 
% our model even achieve better perfomance when compared with large xxx model
% Compared with the large pre-trained grounded language-image model (\ie, GLIP), our model only has 1/8 number of parameters, significantly reducing the computational cost ( FLOPs xx \vs xx) and speeding up the inference process (xx images/s). Even with xx, our model ou
% improving the performance on three benchmark dataset 
% \vspace{-0.5cm}
\end{abstract}

% Inspired by \cite{decoupled1,decoupled2,decoupled3,decoupled4}, we alleviate the inclusion of unknown objects on the known detecting performance by decoupling the detecting process. To alleviate the confusion between the categories and location of the same object caused by the decoupling structure, we propose the cascade structure where foreground localization can be protected from category knowledge, and the identification process can utilize the localization information.

\section{Introduction}
Open-world object detection (OWOD) is a more practical detection problem in computer vision,  facilitating the development of object detection (OD) \cite{detection1,detection2,detection3,detection4,detection5,detection6,detection7,fasterrcnn,ddetr,detr,zou2019object} in the real world. Within the OWOD paradigm, the model’s lifespan is pushed by an iterative learning process. At each episode, the model trained on the data with known object annotations is expected to detect known objects and unknown objects. Human annotators then label a few of these tagged unknown classes of interest gradually. The model given these newly-added annotations continues incrementally updating its knowledge without retraining from scratch. 

In the existing works \cite{ORE, owdetr, OCPL, two-branch}, the models are expected to know about the open world through several datasets \cite{mscoco,voc} with tiny scales. However, the annotations of these datasets are too few to provide adequate object attributes for the model. It is difficult for the model to achieve the ideal goal through these datasets. Humans' ability to recognize objects they have not seen before largely depends on their brains' knowledge base. Inspired by how humans face the open world, we propose searching for a ``Brain'' to assist the learning process of the open-world detector.

Fortunately, the large pre-trained language-image models (VL) \cite{glip,glipv2,mdetr,clip,detclip} could fill this role. They have rich knowledge of the open world due to countless millions of parameters, open datasets, and training costs. However, their detection could not leave the participation of text prompts. Before detecting, the text prompt of all objects must be pre-listed so that they can be detected, and the object whose text prompt was not listed could not be detected.  However, there are countless categories of objects in the open world, and the variety continues to grow. Therefore, the VL could not ideally detect all categories in the open world. In addition, their detection speed is also a criticism due to the following question. $i)$ The huge number of parameters and FLOPs. $ii)$ The large pre-trained grounded language-image model could only infer with several text prompts for the detecting performance, so they must infer many times when the number of prompts is large.

In this paper, we propose to leverage the large pre-trained grounded language-image models to help our detector know about the open world. Our model draws on the knowledge of the large pre-trained grounded language-image model and utilize that to understand the world better so that it recognizes unseen objects beyond the ``Brain''. In addition, our detector identifies all objects beyond its known sets as `unknown' instead of the exact categories. Thus, the text prompt could not limit our detector. During the use of the ``Brain'', we simply leverage its own knowledge to generate unknown labels to assist the training process. For a fair comparison, we use the same dataset and training costs as the existing methods \cite{owdetr,OCPL,ORE,two-branch}, and no extra augmentations are utilized during the training phase. In addition, we do the following works to address the problems encountered while using the ``Brain''.

(\uppercase\expandafter{\romannumeral1}) To prevent the model from only learning unknown classes in the ``unknown labels'', we set a box score prediction branch to help the detector explore unseen objects beyond the large pre-trained grounded language-image model. The box score branch leverages the knowledge of known objects in the annotation and unknown objects in the large pre-trained language-image model to learn the difference between foreground and background. We then leverage the predicted box score to select pseudo-unknown labels from the remaining regression boxes after the matching process of each training iteration. Therefore, our detector could continuously evolve and learn the unknown beyond the large pre-trained grounded language-image model.

(\uppercase\expandafter{\romannumeral2}) It is non-trivial to directly leverage the generated labels because their quality cannot be guaranteed. The performance of detecting known objects is also crucial for OWOD. Through the experiments, we investigate that the direct use of generated labels dramatically affects the model's learning ability of the original annotations. The model's performance on detecting known objects is damaged substantially. To alleviate this, we propose the down-weight training loss for the generated labels, which utilizes the labels' object confidence from the large pre-trained grounded language-image model to generate soft labels \cite{soft1,soft2,soft3} and reduce the weight of unknown loss in the total loss during training. 

(\uppercase\expandafter{\romannumeral3}) The presence of objects with highly similar features to known classes within the ``unknown objects'' can greatly affect the process of open-world object identification. This issue impacts not only the identification process but also the localization process for models that use coupled information for both tasks. Therefore, we propose decoupling the detecting process. Meanwhile, to alleviate the confusion between the category and location of the same object inevitably caused by the decoupling structure, we propose the cascade structure that decouples the detecting process via  two decoders and connects the two decoders via a cascade manner. In this structure, foreground localization can be protected from category knowledge because the loss of identification is diluted by the latter decoder. Moreover, the identification process can utilize the localization information because it leverages the former decoder's output embeddings as the input queries.

Extensive experiments demonstrate that our model outperforms the large pre-trained grounded language-image model for open-world object detection, though it only consumes little training data and computing resources. Our contributions can be summarized fourfold: 
\begin{itemize}\setlength{\itemsep}{-2pt}
\item[$\bullet$] We firstly propose leveraging the large pre-trained grounded language-image model as the ``Brain'' to assist the learning process of the open-world detector by simply generating unknown labels. With the help of the box score branch, the open-world detector could continuously evolve and learn new knowledge beyond the large pre-trained grounded language-image model.
\item[$\bullet$]To mitigate the effect of generated labels on the detection performance of known objects, we propose the down-weight training loss function for the detector's learning process of unknown labels.
\item[$\bullet$] We propose a cascade decoupled detection transformer structure to alleviate the influence caused by unknown objects on detecting known objects.
\item[$\bullet$] Our extensive experiments on two popular benchmarks demonstrate the effectiveness of our model. It outperforms all state-of-the-art methods for OWOD and IOD. Remarkably, our model exceeds the assistant large pre-trained grounded language-image model for OWOD by using minimal training resource.
\end{itemize}

%-------------------------------------------------------------------------
\section{Related Works}
\noindent\textbf{Large pre-trained language-image models:} Recently, inspired by the success of vision-language(VL) pre-training methods \cite{clip} and their good zero-shot ability, some works \cite{vild,glip,mdetr,glipv2,detclip} attempted to perform zero-shot detection on a larger range of domains by using pre-trained vision language models. ViLD \cite{vild} proposed a zero-shot detection method to distill knowledge from a pre-trained vision language image classification model. GLIP \cite{glip} tried to align region and language features using a dot-product operation and could be trained end-to-end on grounding and detection data. MDETR \cite{mdetr} proposed an end-to-end modulated detector that detects objects in an image conditioned on a raw text query, like a caption or a question. DETCLIP \cite{detclip} proposed a paralleled visual-concept pre-training method for open-world detection by resorting to knowledge enrichment from a designed concept dictionary. 
\par
% \noindent\textbf{Semi-Supervised Learning For Object Detection:} In this area, there are two dominant approaches, the consistency methods \cite{ssl1,ssl2} and pseudo-label methods \cite{simplessl,etessl,ssl4,ssl3}, respectively. STAC \cite{simplessl} deploys highly confident pseudo labels of localized objects from an unlabeled image and updates the model by enforcing consistency via strong augmentations. Xu $et$ $al$.\cite{etessl} proposed an end-to-end pseudo-labeling framework  to avoid the complicated training process and also achieves better performance. Liu $et$ $al$.\cite{ssl4} imporved the pseudo-label generation model via teacher-student mutual learning regimen and addresses the crucial imbalance issue in generated pseudo-labels .
% \par
\noindent\textbf{Open-World Object Detection (OWOD):} Joseph \etal \cite{ORE} introduced OWOD task and ORE which adapted the faster-RCNN model with feature-space contrastive clustering, an RPN-based unknown detector, and an Energy Based Unknown Identifier (EBUI) for the OWOD objective. Recently, various works \cite{OCPL,two-branch,Revisiting,UC-OWOD,MViTs} attempted to extend ORE. OCPL \cite{OCPL} was proposed to learn the discriminative embeddings of known classes in the feature space to minimize the overlapping distributions of known and unknown classes. Wu \etal \cite{two-branch} proposed a two-branch objectness-centric open-world object detection framework consisting of the bias-guided detector and the objectness-centric calibrator. OW-DETR \cite{owdetr}, a transformer-based method, was proposed to utilize the pseudo-labeling scheme to supervise unknown object detection, where unmatched object proposals with high backbone activation are selected as unknown objects. 

\begin{figure*}[htbp]
  \setlength{\abovecaptionskip}{0.15cm}
  \centering
  \includegraphics[width = \textwidth]{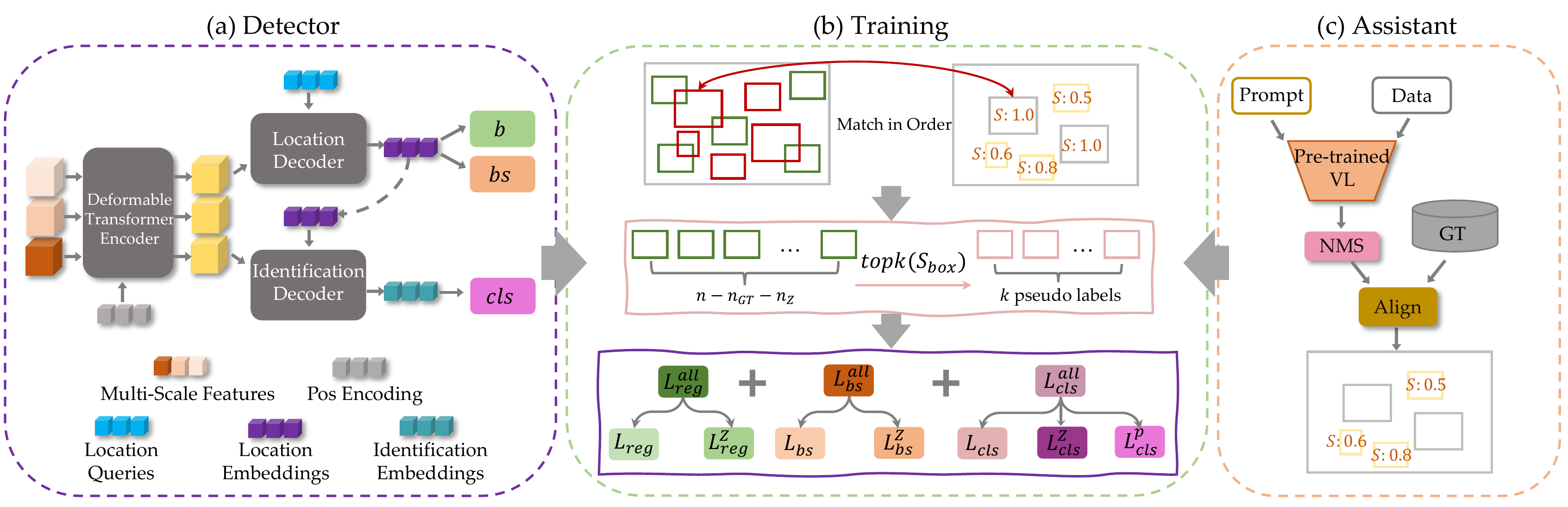}
  \caption{Overall scheme of the proposed framework. (a) illustrates the overall structure of the cascade open-world object detector. (b) exhibits the process of matching, pseudo-labeling, and training. (c) describes the assistant procedure.}
  \label{fig:Figure3}
  \vspace{-0.5cm}
\end{figure*}

\section{Problem Formulation}
$\mathcal{K}^{t}=\{1,2, \ldots, C\}$ denote the set of known object classes and $\mathcal{U}^{t}=\{C+1, \ldots\}$ denote the unknown classes which might be encountered at the test time, at the time $t$. We labeled the known object categories $\mathcal{K}^{t}$ in the dataset $\mathcal{D}^{t}=\{\mathcal{J}^{t}, \mathcal{L}^{t}\}$ where $\mathcal{J}^{t}$ denotes the input images and $\mathcal{L}^{t}$ denotes the corresponding labels at time $t$. The training image set consists of $M$ images $\mathcal{J}^{t}=\{i_{1}, i_{2}, \ldots, i_{M}\}$ and corresponding labels $\mathcal{L}^{t}=\{\ell_{1}, \ell_{2}, \ldots, \ell_{M}\}$. Each $\ell_{i}=\{\mathcal{T}_{1}, \mathcal{T}_{2}, \ldots, \mathcal{T}_{N}\}$ denotes a set of $N$ object instances with their class labels $c_{n} \subset \mathcal{K}^{t}$ and locations, $\{ x_{n}, y_{n}, w_{n}, h_{n}\}$ denote the bounding box center coordinates, width and height respectively. 

The artificial assumptions and restrictions in closed-set object detection are removed in Open-World Object Detection. It aligns object detection tasks more with real life. It requires the trained model $\mathcal{M}_{t}$ to detect the previously encountered known classes $C$ and identify an unseen class instance as belonging to the unknown class. In addition, it requires the object detector to be capable of incremental updates for new knowledge, and this cycle continues over the detector’s lifespan. In the incremental updating phase, the unknown instances identified by $\mathcal{M}_{t}$ are annotated manually. Along with their corresponding training examples, they update $\mathcal{D}^{t}$ to $\mathcal{D}^{t+1}$ and $\mathcal{K}^{t}$ to $\mathcal{K}^{t+1}=\{1,2,\ldots, C,\ldots, C+\text{n}\}$. The model adds the $n$ new classes to known classes and updates itself to $\mathcal{M}_{t+1}$ without retraining from scratch on the whole dataset $\mathcal{D}^{t+1}$.

\section{Proposed method}
This section elaborates on the proposed framework in detail. In Sec.\ref{over scheme}, we illustrate the overall scheme of our proposed framework. In Sec.\ref{detector}, \ref{assistant} and \ref{matching}, we introduce the open-world object detector, the assistant, and the matching and pseudo-labeling procedure. Then, we describe the down-weight training strategy and inference phrase in Sec.\ref{training} and \ref{inference}.

\subsection{Overall Scheme}\label{over scheme}
Fig.\ref{fig:Figure3} illustrates the overall scheme of our framework. For a given image $x \in \mathbb{R}^{H \times W \times3}$, it is first sent into the open-world detector and assistant simultaneously. The detector leverages the visual features of the input to predict the localization, box score, and classification. The assistant utilizes the large pre-trained grounded language-image model to mine more open-world information from the inputs. In the training phase, we match the prediction and open-world labels according to the regression loss, classification, and label confidence. After matching, the pseudo labels are selected according to the predicted box score. Then all labels are leveraged to train the open-world detector by the down-weight training loss function.

\subsection{Cascade Open-World Object Detector}\label{detector}
% Detection transformer \cite{detr,ddetr,misra2021end,beal2020toward,li2022exploring,dai2021dynamic} leverages the object queries to detect object instances, where each object query represents an object instance. Inspired by the literature \cite{decoupled1,decoupled2,decoupled3,decoupled4}, we investigate that decreasing the mission complexity alleviates the model's confusion for objects. Therefore, we propose to decouple the detecting process via  two decoders. To alleviate the confusion of categories and positions of the same object caused by the decoupling structure. We propose to connect them in a cascade manner. In this structure, foreground localization can be protected from category knowledge, and the identification process can utilize the localization information.  In addition, when new categories are introduced at each episode, based on an exemplar replay-based finetuning to alleviate the catastrophic forgetting of learned classes and the finetuning by using a balanced set of exemplars stored for all known classes, our detector could evolve continuously during its lifespan.

The open-world detector first uses a hierarchical feature extraction backbone to extract multi-scale features $\mathrm{Z}_{i} \in \mathbb{\mathbb { R }} ^{ \frac{\mathrm{H}}{4 \times i^{2}} \times \frac{w}{4 \times 2^{i}} \times 2^{i} C_{s}}, i=1,2,3$. The feature maps ${Z}_{i}$ are projected from dimension $C_s$ to dimension $C_d$ by using 1×1 convolution and concatenated to $N_s$ vectors with $C_d$ dimensions after flattening out. Afterwards, along with supplement positional encoding $P \in \mathbb{R}^{N_{s} \times C_{d}}$, the multi-scale features are sent into the deformable transformer encoder to encode semantic features. The encoded semantic features are acquired and sent into the localization decoder together with a set of $N$ learnable location queries. Aided by interleaved cross-attention and self-attention modules, the localization decoder transforms the location queries $Q_{Location} \in \mathbb{R}^{N \times D}$ to a set of N location query embeddings $E_{Location} \in \mathbb{R}^{N \times D}$.  Meanwhile, the $E_{Location} $ are used as class queries and sent into the identification decoder together with the $M$ again. The identification decoder transforms the class queries to $N$ class query embeddings $E_{class}$ that correspond to the location query embeddings. The operation of cascade decoders is expressed as follows:
\begin{equation}\label{eq5}
E_{Location}=F_{LD}(F_E(\varnothing(x), P), Q_{Location}, R),
\end{equation}
\begin{equation}\label{eq6}
E_{Class}=F_{ID}(F_E(\varnothing(x), P), E_{Location}, R).
\end{equation}
where $F_{LD}(\cdot)$ and $F_{ID}(\cdot)$ denote the localization and identification decoder. $F_{E}(\cdot)$ is the encoder and $\varnothing(\cdot)$ is the backbone. $R$ represents the reference points and $x$ denotes the input image. $Q_{Class}$ stands for the class queries. Eventually, the $E_{Class}$ are then sent into the classification branch to predict the category $cls \in [0,1]^{N_{obj}}$. The $E_{Location} $ are then input to the regression and box score branch to locate N foreground bounding boxes $b \in [0,1]^{4}$ and predict the box score $bs \in [0,1]$.

% In this decoupling structure, foreground localization can be protected from category knowledge, and the identification process can utilize the localization information. Therefore, we alleviate the infusion of the unknown objects on the detecting performance of known objects and the confusion between the category and location of the same objects.
In addition, when new categories are introduced at each episode, based on an exemplar replay-based finetuning to alleviate the catastrophic forgetting of learned classes and the finetuning by using a balanced set of exemplars stored for all known classes, our detector could continuously learn during its lifespan.

\subsection{Assistance from the large pre-trained model}\label{assistant}
To help the detector know about the open world, we simply generate labels for the given data during the training. In the assisting phase, the assistant leverages a large pre-trained grounded language-image model and a text prompt that contains object categories as many as possible to help the detector mine information from $x$. In this paper, we utilize GLIP \cite{glip} and categories of LVIS \cite{lvis} as the large pre-trained grounded language-image model and text prompts, respectively. The generated labels are first processed through the NMS produce. Then we align the generated labels and ground truth by the align module, where we align the generated labels to the given data annotation space and exclude the generated labels of the known set. In addition, the identification confidence of the unknown labels from the large pre-trained grounded language-image model is reserved for the following training process. 

\subsection{Matching and Evolving}\label{matching}
When matching the labels and prediction, we consider the box loss between the prediction box and ground truth, prediction class score, and confidence of labels. After matching, we leverage the box score branch to help the model learn novel objects beyond the ground truth and the large pre-trained grounded language-image model during training via selecting pseudo labels. We denote by $\hat{y}$ the ground truth and labels from the large pre-trained grounded language-image model, and $y={\{y_i\}}_{i=1}^N$ the set of $N$ predictions. For finding the best bipartite matching between them, a permutation of $N$ elements $\sigma \in \mathfrak{S}_{N}$ with the lowest cost is searched for as follows:
\begin{equation}
\hat{\omega}=\underset{\omega \in \mathfrak{S}_N}{\arg \min } \sum_i^N \mathcal{L}_{\operatorname{match}}(\boldsymbol{\hat{y}_i}, \boldsymbol{y_{\omega(i)}}),
\end{equation}
where $\mathcal{L}_{\text {match }}\left(\hat{y}_i, y_{\omega(i)}\right)$ is the pair-wise matching cost between $\hat{y}_i$ which represents ground truth or labels from the large pre-trained grounded language-image model and a prediction with index $\omega(i)$, shown as Equation.\ref{match}. Inspired by \cite{ddetr,detr,mdetr,MViTs}, we choose the Hungarian algorithm as the optimal assignment.
\begin{equation}\label{match}
\mathcal{L}_{\operatorname{match}}(\hat{y}_i, {y}_{\omega(i)}) = L_r(\boldsymbol{\hat{b}_i}, \boldsymbol{{b}_{\omega(i)}})- \boldsymbol{cls}_{\omega(i)}(\hat{c}_i)-\boldsymbol{\hat{S}_i},
\end{equation}
where $L_r$ denotes the regression loss, which consists of box loss and GIOU loss \cite{GIOU}. $\hat{b}$ and $b$ represent the label box and prediction box, respectively. $\hat{S}$ is the confidence of the label. Then, the pseudo is selected as follows:
\begin{equation}
l_p =\operatorname{Topk}(\{\boldsymbol{bs}_i\}_{i \notin \widehat{\omega}}),
\end{equation}
where $bs$ denotes the prediction box score. The pseudo-labels could prevent the model from falling entirely into the knowledge of the large pre-trained grounded language-image model and help it know unseen objects beyond the large pre-trained grounded language-image model. With the help of pseudo labels, the ability of the detector to explore unseen objects is stored so that it could continually evolve.

\subsection{Down-Weight Training Strategy}\label{training}
The inclusion of generated labels inevitably influences the model's learning of the known set. Because the quality of generated labels could not be guaranteed, and they increase the difficulty of detector learning. Therefore, we propose the down-weight training strategy, which leverages the generated identification confidence to generate soft labels \cite{soft1,soft2,soft3} and down-weight the unknown training loss and train the detector in an end-to-end manner as shown in Fig.\ref{fig:Figure3} (b). The training loss function is as follows:
\begin{equation}
L=L_{r}+L_{bs}+L_{cls}+L_{r}^z+L_{bs}^z+L_{cls}^z+L_{cls}^p, 
\end{equation}

where the $L_{r}$ uses the common regression loss which consists of box and GIOU loss \cite{GIOU}. $L_{bs}$ and $L_{cls}$ represent the box score and classification loss, respectively. They all leverage the common sigmoid focal loss \cite{focal}. In addition, the $L_{r}^z$, $L_{bs}^z$, $L_{cls}^z$ and $L_{cls}^p$ all utilize correspondingly down-weight loss function we propose, the formulations are shown as follows:

\begin{footnotesize}
\begin{equation}
L_{r}^z=\frac{1}{|\boldsymbol{l_z}|} \sum_{i=1}^{N_q} \mathbbm{1}_{\{i \in \boldsymbol{l_z}\}}\hat{\boldsymbol{S}}_{\hat{\omega}(i)}[\|\boldsymbol{b}_i-\hat{\boldsymbol{b}}_{\hat{\omega}(i)}\|_1+1-\mathcal{G}(\boldsymbol{b}_i, \hat{\boldsymbol{b}}_{\hat{\omega}(i)})],
\end{equation}
\end{footnotesize}

\begin{footnotesize}
\begin{equation}
L_{bs}^z=\frac{1}{\sum_{i=1}^{N_q} \mathbbm{1}_{\{i \in \boldsymbol{l_z}\}}\|\boldsymbol{bs}_i\|_1} \sum_{i=1}^{N_q}\mathbbm{1}_{\{i \in \boldsymbol{l_z}\}}[l_{sf}(\boldsymbol{bs}_i, \hat{\boldsymbol{S}}_{\hat{\boldsymbol{\omega}}(i)})],
\end{equation}
\end{footnotesize}

\begin{footnotesize}
\begin{equation}
L_{cls}^z=\frac{1}{\sum_{i=1}^{N_q} \mathbbm{1}_{\{i \in \boldsymbol{l_z}\}}\|\boldsymbol{cls}_i\|_1} \sum_{i=1}^{N_q}\mathbbm{1}_{\{i \in \boldsymbol{l_z}\}}[l_{sf}(\boldsymbol{cls}_i, \hat{\boldsymbol{S}}_{\hat{\boldsymbol{\omega}}(i)})],
\end{equation}
\end{footnotesize}

\begin{footnotesize}
\begin{equation}
L_{cls}^p=\frac{1}{\sum_{i=1}^{N_q} \mathbbm{1}_{\{i \in \boldsymbol{l_p}\}}\|\boldsymbol{cls}_i\|_1} \sum_{i=1}^{N_q}\mathbbm{1}_{\{i \in \boldsymbol{l_p}\}}[l_{sf}(\boldsymbol{cls}_i, \boldsymbol{bs}_i)],
\end{equation}
\end{footnotesize}

where $N_q$ represents the number of queries, $\hat{\omega}(i)$ represents the index of the label corresponding to the prediction. $l_{sf}$ denotes the sigmoid focal loss function. $\mathcal{G}(\cdot)$ represents the GIOU loss function. $\hat{S}$ is the confidence of labels. $b$, $bs$, and $cls$ are the prediction box, box score, and classification score, respectively.

\subsection{Inference}\label{inference}
 Our detector only utilizes the visual features of the inputs to detect open-world objects without any information from the other modalities. The inference process is to composite the detector output to form open-world object instances. Formally, the $i$-th output prediction is generated as $<b_i,\ bs_i,\ cls_i>$. According to the formulation : $s_i, l_i = max(cls_i)$, the result is acquired as $<l_i,\ s_i,\ b_i>$, where $l$ is the category, $s$ denotes the confidence, and $b$ represents the predicted bounding boxes.

\begin{table*}[]

\centering

\resizebox{\textwidth}{!}{
\begin{tabular}{c|c| ccc|ccc| ccc |ccc }\toprule
\multicolumn{1}{c|}{}&Task IDs $\rightarrow$ &&&& \multicolumn{3}{c|}{Task 1}& \multicolumn{3}{c|}{Task 2} & \multicolumn{3}{c}{Task 3}   \\\midrule
\multicolumn{1}{c|}{}  &&&&Inference&\cellcolor[HTML]{cff5ce}Train& \multicolumn{1}{c}{ \cellcolor[HTML]{fcfce0}Unknown}  & \cellcolor[HTML]{e4e5fa}Known & \cellcolor[HTML]{cff5ce}Train& \multicolumn{1}{c}{ \cellcolor[HTML]{fcfce0}Unknown} & \multicolumn{1}{c|}{\cellcolor[HTML]{e4e5fa}Known} & \cellcolor[HTML]{cff5ce}Train& \multicolumn{1}{c}{ \cellcolor[HTML]{fcfce0}Unknown} & \multicolumn{1}{c}{\cellcolor[HTML]{e4e5fa}Known}    \\

 \multicolumn{1}{c|}{}& &\multicolumn{1}{c}{Param\#}  & \multicolumn{1}{c}{FLOPs} & \multicolumn{1}{c|}{Rate} &\cellcolor[HTML]{cff5ce}Data &\cellcolor[HTML]{fcfce0} Recall&\cellcolor[HTML]{e4e5fa}mAP($\uparrow$) &\cellcolor[HTML]{cff5ce}Data &\cellcolor[HTML]{fcfce0}Recall& \multicolumn{1}{c|}{\cellcolor[HTML]{e4e5fa}mAP($\uparrow$)}&\cellcolor[HTML]{cff5ce}Data&\cellcolor[HTML]{fcfce0}Recall  &\multicolumn{1}{c}{\cellcolor[HTML]{e4e5fa}mAP($\uparrow$)}  \\

\multicolumn{1}{c|}{\multirow{-3}{*}{SPLIT}}&\multirow{-3}{*}{Metrics $\rightarrow$} &&&(s/img)&\cellcolor[HTML]{cff5ce}(imgs) & \multirow{-1}{*}{\cellcolor[HTML]{fcfce0}($\uparrow$)} & \cellcolor[HTML]{e4e5fa}Both &\cellcolor[HTML]{cff5ce}(imgs)& \multirow{-1}{*}{\cellcolor[HTML]{fcfce0}($\uparrow$)}  &\cellcolor[HTML]{e4e5fa}Both&\cellcolor[HTML]{cff5ce}(imgs) & \multirow{-1}{*}{\cellcolor[HTML]{fcfce0}($\uparrow$)} &\cellcolor[HTML]{e4e5fa}Both \\ \midrule
% \multicolumn{16}{c}{\textbf{\large{OWOD SPLIT}}} \\ \midrule
 \multirow{2}{*}{OWOD} & GLIP&321.9M&965GMac& 9.22&64M&37.0   &42.0 &64M& 35.5  & 32.4 &64M & 34.9 &28.2\\ 
 \multirow{1}{*}{}&\cellcolor[HTML]{efefef}\textbf{Ours} & \cellcolor[HTML]{efefef}\textbf{42.9M}&\cellcolor[HTML]{efefef}\textbf{212GMac}& \cellcolor[HTML]{efefef}\textbf{0.08}& \cellcolor[HTML]{efefef}\textbf{0.83M}    & \cellcolor[HTML]{efefef}\textbf{39.0} &\cellcolor[HTML]{efefef}\textbf{56.8} & \cellcolor[HTML]{efefef}\textbf{0.35M}&\cellcolor[HTML]{efefef}\textbf{36.7} & \cellcolor[HTML]{efefef}\textbf{40.3}  &\cellcolor[HTML]{efefef}\textbf{0.27M} &\cellcolor[HTML]{efefef}\textbf{36.1} &\cellcolor[HTML]{efefef}\textbf{30.1}\\ \midrule
% \multicolumn{16}{c}{\textbf{\large{MS-COCO SPLIT}}} \\ \midrule
 \multirow{2}{*}{MS-COCO} & GLIP  &321.9M&965GMac& 9.22&64M&52.6    &46.5 &64M& 54.5   & 37.7 &64M & 53.3 &35.2\\ 
 \multirow{1}{*}{}&\cellcolor[HTML]{efefef}\textbf{Ours} &  \cellcolor[HTML]{efefef}\textbf{42.9M}&\cellcolor[HTML]{efefef}\textbf{212GMac}& \cellcolor[HTML]{efefef}\textbf{0.08}& \cellcolor[HTML]{efefef}\textbf{4.0M} &\cellcolor[HTML]{efefef}\textbf{60.9}    &\cellcolor[HTML]{efefef}\textbf{69.4} & \cellcolor[HTML]{efefef}\textbf{0.36M}&\cellcolor[HTML]{efefef}\textbf{60.0} & \cellcolor[HTML]{efefef}\textbf{44.4}  &\cellcolor[HTML]{efefef}\textbf{0.37M} &\cellcolor[HTML]{efefef}\textbf{58.6} &\cellcolor[HTML]{efefef}\textbf{40.1}
 \\\bottomrule
\end{tabular}}
\setlength{\abovecaptionskip}{0.15cm}
\caption{Comparison with the large pre-trained grounded language-image model for Open-World object detection on OWOD and MS-COCO split. \texttt{Train Data} represents the total number of image inputs (\texttt{iters} $\times$ $\texttt{batch}_\texttt{size}$) during the training time. For GLIP, the prediction beyond the known categories is set to unknown. } 
\label{table1}
\end{table*}

\begin{figure*}[htbp]
  \setlength{\abovecaptionskip}{0.15cm}
  \centering
  \includegraphics[width = 0.95\textwidth]{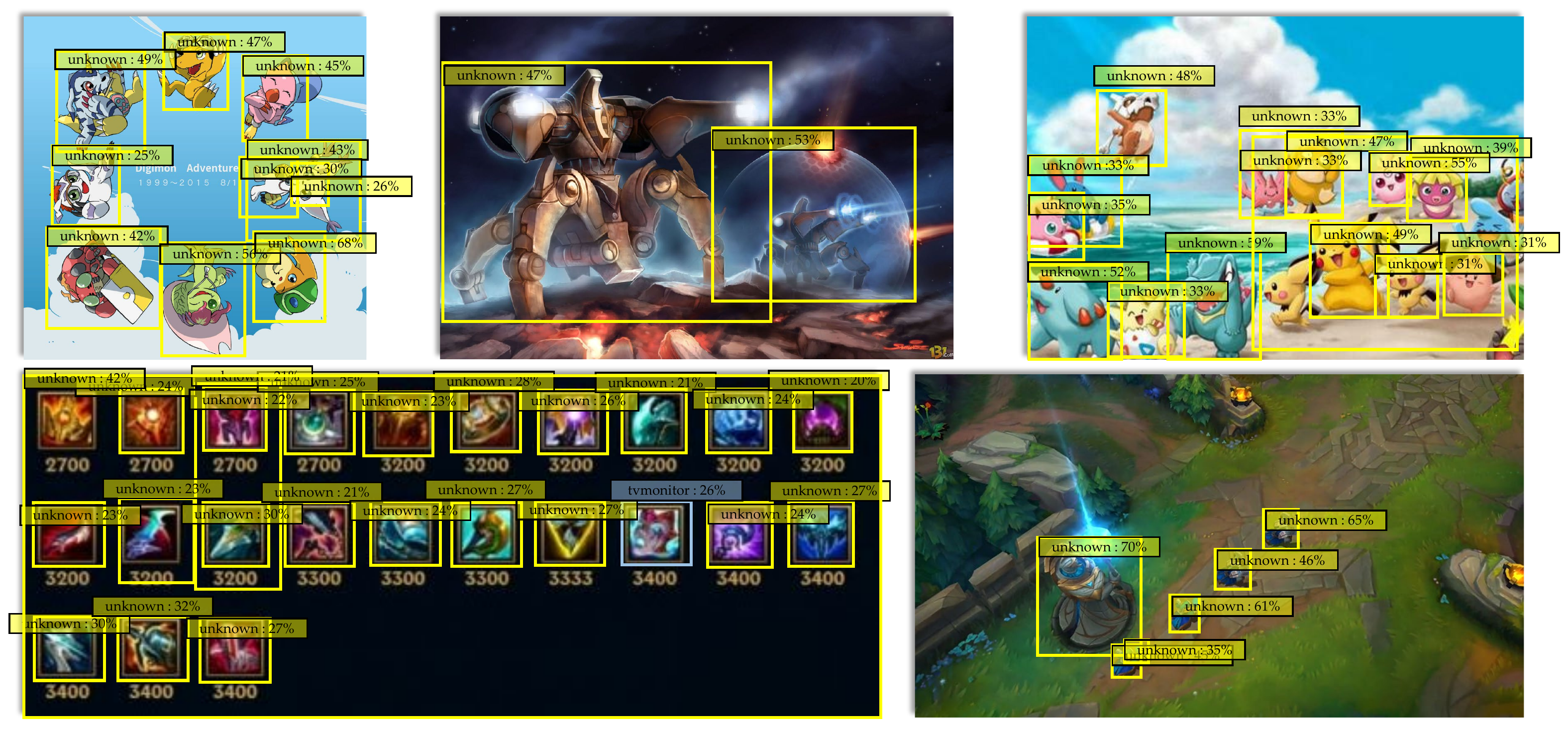}
  \caption{Visualization results out of the LVIS \cite{lvis} text prompts. Visualization results are based on the setting of Task.1. Our model can detect the unknown objects in {\textcolor[RGB]{255,215,0}{Yellow}} boxes beyond the unknown labels from GLIP \cite{glip} and LVIS \cite{lvis} text prompts.}
  \label{fig:Figure3}
  \vspace{-0.5cm}
\end{figure*}
% \vspace{-0.3cm}

\begin{table*}[htbp]
\renewcommand\arraystretch{1.25}
\centering
\resizebox{\textwidth}{!}{
\begin{tabular}{l| cc| cccc |cccc |ccc}\toprule
\multicolumn{1}{c|}{Task IDs $\rightarrow$}& \multicolumn{2}{c|}{Task 1}& \multicolumn{4}{c|}{Task 2} & \multicolumn{4}{c|}{Task 3}  & \multicolumn{3}{c}{Task 4} \\\midrule
\multicolumn{1}{c|}{}   & \multicolumn{1}{c}{ \cellcolor[HTML]{fcfce0}Unknown}  & \cellcolor[HTML]{e4e5fa}Known &  \multicolumn{1}{c}{ \cellcolor[HTML]{fcfce0}Unknown} & \multicolumn{3}{c|}{\cellcolor[HTML]{e4e5fa}Known} & \multicolumn{1}{c}{ \cellcolor[HTML]{fcfce0}Unknown} & \multicolumn{3}{c|}{\cellcolor[HTML]{e4e5fa}Known}  &\multicolumn{3}{c}{\cellcolor[HTML]{e4e5fa}Known}  \\
\multicolumn{1}{l|}{}    & \cellcolor[HTML]{fcfce0} Recall&  \cellcolor[HTML]{e4e5fa}mAP($\uparrow$) & \cellcolor[HTML]{fcfce0}Recall& \multicolumn{3}{c|}{\cellcolor[HTML]{e4e5fa}mAP($\uparrow$)}&\cellcolor[HTML]{fcfce0}Recall  &\multicolumn{3}{c|}{\cellcolor[HTML]{e4e5fa}mAP($\uparrow$)}   &  \multicolumn{3}{c}{\cellcolor[HTML]{e4e5fa}mAP($\uparrow$)}\\

\multicolumn{1}{c|}{\multirow{-3}{*}{Metrics $\rightarrow$}}   & \multirow{-1}{*}{\cellcolor[HTML]{fcfce0}($\uparrow$)}& \cellcolor[HTML]{e4e5fa}Current & \multirow{-1}{*}{\cellcolor[HTML]{fcfce0}($\uparrow$)}   &\cellcolor[HTML]{e4e5fa}Previously & \cellcolor[HTML]{e4e5fa}Current &\cellcolor[HTML]{e4e5fa}Both & \multirow{-1}{*}{\cellcolor[HTML]{fcfce0}($\uparrow$)}  &\cellcolor[HTML]{e4e5fa}Previously & \cellcolor[HTML]{e4e5fa}Current &\cellcolor[HTML]{e4e5fa}Both &\cellcolor[HTML]{e4e5fa}Previously & \cellcolor[HTML]{e4e5fa}Current &\cellcolor[HTML]{e4e5fa}Both \\ \midrule

UC-OWOD\cite{UC-OWOD} &\cellcolor[HTML]{fcfce0}2.4  & 50.7 & \cellcolor[HTML]{fcfce0}3.4  & 33.1& 30.5 & 31.8 &\cellcolor[HTML]{fcfce0}8.7  &28.8&16.3&24.6&25.6&15.9&23.2 \\
ORE-EBUI\cite{ORE}& \cellcolor[HTML]{fcfce0}4.9  & 56.0   &\cellcolor[HTML]{fcfce0}2.9  & 52.7 & 26.0  & 39.4  & \cellcolor[HTML]{fcfce0}3.9   & 38.2  & 12.7 & 29.7  & 29.6   & 12.4  & 25.3  \\
OW-DETR\cite{owdetr}  &\cellcolor[HTML]{fcfce0}7.5 &  59.2 & \cellcolor[HTML]{fcfce0}6.2 & 53.6  & 33.5   & 42.9  & \cellcolor[HTML]{fcfce0}5.7  & 38.3 & 15.8 & 30.8  & 31.4 & 17.1 & 27.8  \\ 
OCPL\cite{OCPL} &\cellcolor[HTML]{fcfce0}8.3  & 56.6 &\cellcolor[HTML]{fcfce0}7.7  & 50.6    & 27.5 & 39.1 &\cellcolor[HTML]{fcfce0}11.9  &38.7&14.7&30.7&30.7&14.4 &26.7\\
2B-OCD\cite{two-branch} &\cellcolor[HTML]{fcfce0}12.1  & 56.4 & \cellcolor[HTML]{fcfce0}9.4  & 51.6& 25.3 & 38.5 &\cellcolor[HTML]{fcfce0}11.6  & 37.2 & 13.2 & 29.2&30.0&13.3&25.8 \\
 \multirow{1}{*}{\textbf{Ours}}   &\cellcolor[HTML]{fcfce0}\textbf{39.0}    &56.8 & \cellcolor[HTML]{fcfce0}\textbf{36.7} & 52.3 & 28.3  & 40.3  & \cellcolor[HTML]{fcfce0}\textbf{36.1}&  36.9 & 16.4  & 30.1 &31.0 & 14.7  &26.9\\ 

\bottomrule 
\end{tabular}}
\setlength{\abovecaptionskip}{0.15cm}
\caption{State-of-the-art comparison for open-world object detection on OWOD split. The comparison is shown in terms of U-Recall and known class mAP.} 
\label{table2}
\end{table*}

\begin{table*}[htbp]
\renewcommand\arraystretch{1.25}
\centering
\resizebox{\textwidth}{!}{
\begin{tabular}{l| cc| cccc |cccc |ccc}\toprule
\multicolumn{1}{c|}{Task IDs $\rightarrow$}& \multicolumn{2}{c|}{Task 1}& \multicolumn{4}{c|}{Task 2} & \multicolumn{4}{c|}{Task 3}  & \multicolumn{3}{c}{Task 4} \\\midrule
\multicolumn{1}{c|}{}   & \multicolumn{1}{c}{ \cellcolor[HTML]{fcfce0}Unknown}  & \cellcolor[HTML]{e4e5fa}Known &  \multicolumn{1}{c}{ \cellcolor[HTML]{fcfce0}Unknown} & \multicolumn{3}{c|}{\cellcolor[HTML]{e4e5fa}Known} & \multicolumn{1}{c}{ \cellcolor[HTML]{fcfce0}Unknown} & \multicolumn{3}{c|}{\cellcolor[HTML]{e4e5fa}Known}  &\multicolumn{3}{c}{\cellcolor[HTML]{e4e5fa}Known}  \\
\multicolumn{1}{l|}{}    & \cellcolor[HTML]{fcfce0} Recall&  \cellcolor[HTML]{e4e5fa}mAP($\uparrow$) & \cellcolor[HTML]{fcfce0}Recall& \multicolumn{3}{c|}{\cellcolor[HTML]{e4e5fa}mAP($\uparrow$)}&\cellcolor[HTML]{fcfce0}Recall  &\multicolumn{3}{c|}{\cellcolor[HTML]{e4e5fa}mAP($\uparrow$)}   &  \multicolumn{3}{c}{\cellcolor[HTML]{e4e5fa}mAP($\uparrow$)}\\

\multicolumn{1}{c|}{\multirow{-3}{*}{Metrics $\rightarrow$}}   & \multirow{-1}{*}{\cellcolor[HTML]{fcfce0}($\uparrow$)}& \cellcolor[HTML]{e4e5fa}Current & \multirow{-1}{*}{\cellcolor[HTML]{fcfce0}($\uparrow$)}   &\cellcolor[HTML]{e4e5fa}Previously & \cellcolor[HTML]{e4e5fa}Current &\cellcolor[HTML]{e4e5fa}Both & \multirow{-1}{*}{\cellcolor[HTML]{fcfce0}($\uparrow$)}  &\cellcolor[HTML]{e4e5fa}Previously & \cellcolor[HTML]{e4e5fa}Current &\cellcolor[HTML]{e4e5fa}Both &\cellcolor[HTML]{e4e5fa}Previously & \cellcolor[HTML]{e4e5fa}Current &\cellcolor[HTML]{e4e5fa}Both \\ \midrule

ORE-EBUI\cite{ORE}& \cellcolor[HTML]{fcfce0}1.5 & 61.4   & \cellcolor[HTML]{fcfce0}3.9& 56.5 & 26.1  & 40.6  &\cellcolor[HTML]{fcfce0}3.6   & 38.7  & 23.7 & 33.7   & 33.6   & 26.3  & 31.8  \\
OW-DETR\cite{owdetr}  & \cellcolor[HTML]{fcfce0}5.7 &  71.5 & \cellcolor[HTML]{fcfce0}6.2 & 62.8  & 27.5   & 43.8  & \cellcolor[HTML]{fcfce0}6.9& 45.2 & 24.9 & 38.5  & 38.2 & 28.1 & 33.1  \\ 

 \multirow{1}{*}{\textbf{Ours}}   &\cellcolor[HTML]{fcfce0}\textbf{60.9}  &69.4 & \cellcolor[HTML]{fcfce0}\textbf{60.0} &63.8 & 26.9  & 44.4  & \cellcolor[HTML]{fcfce0}\textbf{58.6}&  46.2 &28.0 & 40.1 &41.8 &29.6  &38.7\\ 
\bottomrule 
\end{tabular}}
\setlength{\abovecaptionskip}{0.15cm}
\caption{State-of-the-art comparison for open-world object detection on MS-COCO split. As the code and weights of UC-OWOD \cite{UC-OWOD}, OCPL \cite{OCPL} and 2B-OCD \cite{two-branch} are not publicly available, we cannot get results of them or evaluate them on MS-COCO split. Thus, we only compare our model with ORE \cite{ORE} and OW-DETR \cite{owdetr}.} 
\label{table3}
\vspace{-0.5cm}
\end{table*}

\section{Experiment}
\subsection{Datasets and Metrics}
For a fair comparison, we implement the experiments on two mainstream splits of MS-COCO \cite{mscoco}, and Pascal VOC \cite{voc} dataset. We group the classes into a set of non-overlapping tasks $\left\{T^1, \ldots, T^t, \ldots\right\}$. The class in task $T^c$ only appears in tasks where $t \geq c$. In task $T^c$, classes encountered in $\left\{T^c: c \leq t\right\}$ and $\left\{T^c: c>t\right\}$ are considered as known and unknown classes, respectively. \par
\noindent \textbf{OWOD SPLIT} \cite{ORE} splits the 80 classes of MS-COCO into 4 tasks and selects a training set for each task from the MS-COCO and Pascal VOC training set. Pascal VOC testing and MS-COCO validation set are used for evaluation. \par
\noindent \textbf{MS-COCO SPLIT} \cite{owdetr} mitigates data leakage across tasks in \cite{ORE} and is more challenging. The training and testing data are selected from MS-COCO.\par
\noindent \textbf{Metrics:} 
Following the most commonly used evaluation metric for object detection, we use mean average precision (mAP) to evaluate the known objects. Following OW-DETR \cite{owdetr}, U-Recall which measures the ability of the model to retrieve unknown object instances for OWOD problems, is used as the metric for unknown objects.
\subsection{Implementation Details}
The multi-scale feature extractor consists of a Resnet-50\cite{resnet50} pretrained on ImageNet\cite{imagenet} in a self-supervised\cite{self} manner and a deformable transformer encoder whose number of layers is set to 6. For the two cascade decoders, we all use the deformable transformer decoder, and the number of layers is set to 6, too. We set the number of queries $M=100$, the dimension of the embeddings $D=256$, and the number of pseudo-labels $k=5$. We set GLIP \cite{glip} as the large pre-trained grounded language-image model and categories of LVIS dataset \cite{lvis} as text prompts to assist the training process. For GLIP, we use the GLIP-L \cite{glip} without finetuning on the COCO dataset \cite{mscoco}. It consists of Swin-Large \cite{swin}, text encoder of CLIP \cite{clip}, DyHead \cite{dai2021dynamic}, BERT Layer \cite{bert} and Fusion Module \cite{glip}. For the incremental object detection experiments, we only use our open-world detector without the help of GLIP.

\subsection{Comparison With the Assistant}
The results compared with the large pre-trained grounded language-image model (GLIP) for OWOD problem are shown in Table.\ref{table1}. Regarding the number of parameters and FLOPs, our model is significantly smaller than GLIP. In particular, the inference speed of ours is $115 \times \sim 116 \times$ of GLIP's. Furthermore, the number of GLIP's training data is 64M images, while our model only needs a small amount of data in each task of different splits almost $\frac{1}{237} \times \sim \frac{1}{16} \times$ of its.  Compared with GLIP's U-Recall of 37.0, 35.5, and 34.9 on Task 1, 2, and 3 of OWOD split, ours achieves 39.0, 36.7 and 36.1 in the corresponding tasks, achieving significant absolute gains up to 2.0, 1.2 and 1.2, respectively. In the MS-COCO split, our model achieves more than 5 point improvement in terms of U-Recall. This demonstrates that our model has a better ability to detect unknown and known objects for OWOD.\par
\noindent\textbf{Qualitative Results:} To present the ability to detect unknown objects intuitively, we select several open scenes from games and comics. These scenes contain the object out of the categories of LVIS \cite{lvis} text prompts. visualization results are shown in Fig.\ref{fig:Figure3}. Benefits from the box score branch and pseudo labels, our detector evolves the novel unknown objects beyond the large pre-trained grounded language-image model.
% \vspace{-0.5cm}
\subsection{Comparison With State-of-the-art Methods}
\noindent \textbf{OWOD SPLIT:}
The results compared with the state-of-the-art methods on OWOD split for OWOD problem are shown in Table.\ref{table2}. The ability of our model to detect unknown objects quantified by U-Recall is more than $3 \times$ of those reported in previous state-of-the-art OWOD methods.  Compared with the model 2B-OCD \cite{two-branch} with the highest U-Recall of 12.1, 9.4 and 11.6 on Task 1, 2, and 3, ours achieves 39.0, 36.7, and 36.1 in the corresponding tasks, achieving significant absolute gains up to 26.9, 27.3 and 24.5, respectively. Benefits from the cascade decoder structure and down-weight training loss which mitigate the effect of unknown objects on detecting known objects, our model's performance on known objects is also superior to most state-of-the-art methods. 
\par
\noindent \textbf{MS-COCO SPLIT:}
We report the results on the MS-COCO split in Table.\ref{table3}. MS-COCO split mitigates data leakage across tasks and assigns more data to each Task, while our model receives a more significant boost compared with the OWOD split. Our model's unknown object detection capability, quantified by U-Recall, is almost $10 \times \sim 11 \times$ of those reported in previous state-of-the-art OWOD methods. Compared with OW-DETR's U-Recall of 5.7, 6.2, and 6.9 on Task 1, 2, and 3, ours achieves 60.9, 60.0, and 58.6 in the corresponding tasks, achieving significant absolute gains up to 55.2, 53.8 and 51.7, respectively. This demonstrates that our model has the more powerful ability to retrieve new knowledge and detect unknown objects when faced with more difficult tasks.
\par

\begin{table}[htbp]
\centering

\resizebox{\linewidth}{!}{
\begin{tabular}{c|ccc}
\toprule
Method & 10+10 settings & 15+5 settings & 19+1 settings \\ \midrule
\multicolumn{1}{c|}{ILOD\cite{iod}}        & 63.2  & 65.8 & 68.2 \\
\multicolumn{1}{c|}{Faster ILOD\cite{fasteriod}} & 62.1  & 67.9 & 68.5 \\
\multicolumn{1}{c|}{ORE\cite{ORE}}         & 64.5  & 68.5 & 68.8 \\
\multicolumn{1}{c|}{OW-DETR\cite{owdetr}}     & 65.7  & 69.4 & 70.2 \\ \midrule
\multicolumn{1}{c|}{\textbf{Ours}}   &{\textbf{68.6}} &{\textbf{72.1}} &{\textbf{72.9}}\\ \bottomrule
\end{tabular}}
\setlength{\abovecaptionskip}{0.15cm}
\caption{State-of-the-art comparison for incremental object detection for three different settings on the PASCAL VOC dataset. The comparison is shown in terms of overall mAP. We only use our open-world object detector without assistance.} 
\label{table4}
%\vspace{-0.2cm}
\vspace{-0.5cm}
\end{table}

\begin{table*}[]
\centering
\label{table5}
\resizebox{\textwidth}{!}{
\begin{tabular}{l| cc| cccc |cccc |ccc}\toprule
\multicolumn{1}{c|}{Task IDs $\rightarrow$}& \multicolumn{2}{c|}{Task 1}& \multicolumn{4}{c|}{Task 2} & \multicolumn{4}{c|}{Task 3}  & \multicolumn{3}{c}{Task 4} \\\midrule
\multicolumn{1}{c|}{}   & \multicolumn{1}{c}{ \cellcolor[HTML]{fcfce0}Unknown}  & \cellcolor[HTML]{e4e5fa}Known &  \multicolumn{1}{c}{ \cellcolor[HTML]{fcfce0}Unknown} & \multicolumn{3}{c|}{\cellcolor[HTML]{e4e5fa}Known} & \multicolumn{1}{c}{ \cellcolor[HTML]{fcfce0}Unknown} & \multicolumn{3}{c|}{\cellcolor[HTML]{e4e5fa}Known}  &\multicolumn{3}{c}{\cellcolor[HTML]{e4e5fa}Known}  \\
\multicolumn{1}{l|}{}    & \cellcolor[HTML]{fcfce0} Recall&  \cellcolor[HTML]{e4e5fa}mAP($\uparrow$) & \cellcolor[HTML]{fcfce0}Recall& \multicolumn{3}{c|}{\cellcolor[HTML]{e4e5fa}mAP($\uparrow$)}&\cellcolor[HTML]{fcfce0}Recall  &\multicolumn{3}{c|}{\cellcolor[HTML]{e4e5fa}mAP($\uparrow$)}   &  \multicolumn{3}{c}{\cellcolor[HTML]{e4e5fa}mAP($\uparrow$)}\\

\multicolumn{1}{c|}{\multirow{-3}{*}{Metrics $\rightarrow$}}   & \multirow{-1}{*}{\cellcolor[HTML]{fcfce0}($\uparrow$)}&\cellcolor[HTML]{e4e5fa}Current & \multirow{-1}{*}{\cellcolor[HTML]{fcfce0}($\uparrow$)}   &\cellcolor[HTML]{e4e5fa}Previously &\cellcolor[HTML]{e4e5fa}Current &\cellcolor[HTML]{e4e5fa}Both & \multirow{-1}{*}{\cellcolor[HTML]{fcfce0}($\uparrow$)}  &\cellcolor[HTML]{e4e5fa}Previously & \cellcolor[HTML]{e4e5fa}Current &\cellcolor[HTML]{e4e5fa}Both &\cellcolor[HTML]{e4e5fa}Previously &\cellcolor[HTML]{e4e5fa}Current &\cellcolor[HTML]{e4e5fa}Both \\ \midrule
\texttt{Baseline}      &  39.4         &\cellcolor[HTML]{e4e5fa}41.1    &  36.6       &\cellcolor[HTML]{e4e5fa}27.9&\cellcolor[HTML]{e4e5fa}10.5 &  \cellcolor[HTML]{e4e5fa}19.2     & 36.6       &\cellcolor[HTML]{e4e5fa}18.0&\cellcolor[HTML]{e4e5fa}6.2  &\cellcolor[HTML]{e4e5fa}14.1  &\cellcolor[HTML]{e4e5fa}14.8 &\cellcolor[HTML]{e4e5fa}5.8&  \cellcolor[HTML]{e4e5fa}12.6 \\
\texttt{Baseline} \texttt{+} \texttt{DW}   &  39.2          &\cellcolor[HTML]{e4e5fa}53.5   &    36.3        &\cellcolor[HTML]{e4e5fa}47.5&\cellcolor[HTML]{e4e5fa}22.3 &   \cellcolor[HTML]{e4e5fa}34.9    &  35.9   &\cellcolor[HTML]{e4e5fa}32.0&\cellcolor[HTML]{e4e5fa}11.5    &\cellcolor[HTML]{e4e5fa}25.1   &\cellcolor[HTML]{e4e5fa}25.4 &\cellcolor[HTML]{e4e5fa}9.6  &\cellcolor[HTML]{e4e5fa}22.2  \\
\texttt{Baseline} \texttt{+} \texttt{CS}    &39.1&\cellcolor[HTML]{e4e5fa}51.9    &  36.9     &\cellcolor[HTML]{e4e5fa}46.7&\cellcolor[HTML]{e4e5fa}22.6  &   \cellcolor[HTML]{e4e5fa}34.6    & 36.1        &\cellcolor[HTML]{e4e5fa}32.4& \cellcolor[HTML]{e4e5fa}12.8      & \cellcolor[HTML]{e4e5fa}25.9   &\cellcolor[HTML]{e4e5fa}26.7&\cellcolor[HTML]{e4e5fa}12.4  &\cellcolor[HTML]{e4e5fa}21.8  \\ 

\texttt{Final: Ours}       &   39.0     &\cellcolor[HTML]{e4e5fa}\textbf{56.8}    & 36.7       &\cellcolor[HTML]{e4e5fa}\textbf{52.3} &\cellcolor[HTML]{e4e5fa}\textbf{28.3}     &\cellcolor[HTML]{e4e5fa}\textbf{40.3}    &   36.1   &\cellcolor[HTML]{e4e5fa}\textbf{36.9}&\cellcolor[HTML]{e4e5fa}\textbf{16.4}  &\cellcolor[HTML]{e4e5fa}\textbf{30.1}      &\cellcolor[HTML]{e4e5fa}\textbf{31.0}&\cellcolor[HTML]{e4e5fa}\textbf{14.7}&\cellcolor[HTML]{e4e5fa}\textbf{26.9}\\
\bottomrule
\end{tabular}}
\setlength{\abovecaptionskip}{0.15cm}
\caption{Experiments on ablating each component. \texttt{DW} represents the down-training loss function. When \texttt{DW} is none, we use the same loss function as GT for unknown labels. \texttt{CS} represents the cascade decoupling structure. When \texttt{CS} is none, we leverage the normal decoder structure as DDETR \cite{ddetr}.} 
\label{table5}
\end{table*}

\begin{figure*}[htbp]
  \setlength{\abovecaptionskip}{0.15cm}
  \centering
  \includegraphics[width = \textwidth]{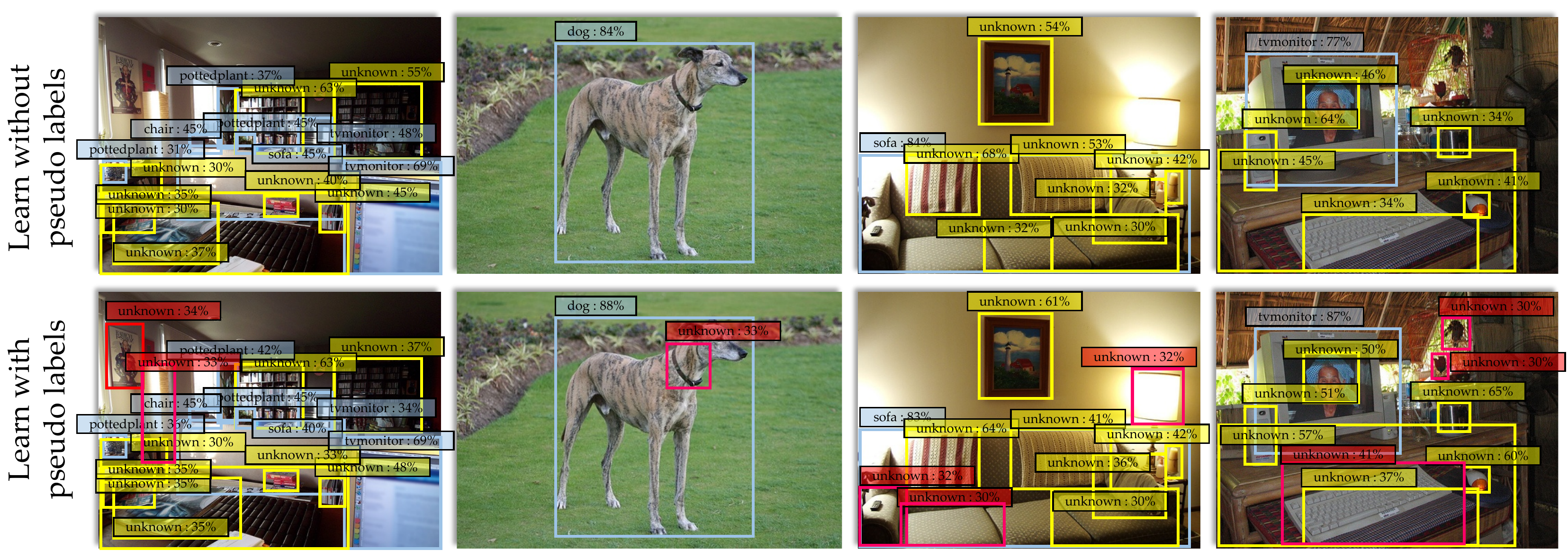}
  \caption{Visualization comparison on ablating pseudo-labeling. Detections are displayed, including \textcolor[rgb]{0.1,0.8,0.9}{Blue} known objects, \textcolor[RGB]{255,215,0}{Yellow} unknown objects detected by both, and \red{Red} unknown objects which are leaked when the box score branch is removed.}
 % $Poster$, $Dog\ Collar$, $Table\ Lamp$, $Remote$ and $Birdcage$.
  \label{fig:Figure4}
  \vspace{-0.5cm}
\end{figure*}

\noindent \textbf{Incremental Object Detection:}
To intuitively present our detector's ability for detecting object instances, we compare it to \cite{iod,fasteriod,ORE,owdetr} on the incremental object detection (IOD) task. We do not use assistance from the large pre-trained grounded language-image model. We evaluate the experiments on three standard settings, where a group of classes (10, 5, and last class) is introduced incrementally to a detector trained on the remaining classes (10, 15, and 19), based on the PASCAL VOC 2007 dataset \cite{voc}. As the results shown in Table.\ref{table4}, our model outperforms the existing method in a great migration on all three settings, indicating the power of cascade detection transformer for IOD.\par

\subsection{Ablation Study}
In this subsection, a set of experiments are designed to clearly understand the contribution of each of the constituent components. We conducted all experiments on the OWOD split.

\noindent\textbf{Ablating Down-Weight Loss and Cascade Structure:} 
To study the contribution of each component, we design ablation experiments in Table.\ref{table5}. In comparison to the baseline, adding the down-weight training loss function significantly improves the performance on detecting known objects and incremental object detecting, achieving significant absolute gains up to more than 10 points. As we have analyzed, the cascade decoupling structure alleviates the inclusion of unknown objects on the known detecting performance and the confusion between the categories and locations of the same objects. It significantly improves the performance of detecting known objects with absolute gains up to more than 10 points, too. What's more, these two combine effectively, improving performance with absolute gains up to 15.7, 24.4, 17.8, 21.1, 18.9, 10.2, 16.0, 16.2, 8.9, and 14.3 in the corresponding tasks without significantly reducing the ability to detect unknown objects. Thus, each component has a critical role to play in open-World object detection. \par

\noindent\textbf{Ablating the Pseudo Labels: }
We also ablate the pseudo labels by controlling the presence or absence of the box score branch. However, owing to the absence of effective metrics, existing metrics cannot demonstrate the difference. In terms of Unknown Recall, they almost have the same performance where the Unknown Recall changes from 39.0, 36.7, 36.1 to 39.1 36.4 36.0 when the box score branch is removed. Therefore, we ablate it according to the several visualization results, as shown in Fig.\ref{fig:Figure4}. When the pseudo labels are removed, the model misdetect many unknown objects. The comparison demonstrates that the pseudo labels can help the detector better know about the world. \par

\section{Conclusions}
% %%\vspace{-0.2cm}
In this paper, we propose to leverage a large pre-trained grounded language-image model to help our detector know about the open world. To prevent the model from learning only unknown
classes in the “unknown labels”, we set a box score prediction branch to help the detector explore unseen objects beyond the large pre-trained grounded language-image model.
To alleviate the problems posed by the co-existence of knowledge of the brain and data annotations, we propose the down-weight training loss and a novel cascade decoupled detection transformer structure. Extensive experiments demonstrate the effectiveness of our model.

\section*{Acknowledgment}
This work is supported by National Nature Science Foundation of China (grant No.61871106), and the Open Project Program Foundation of the Key Laboratory of Opto-Electronics Information Processing, Chinese Academy of Sciences (OEIP-O-202002).

{\small
\bibliographystyle{ieee_fullname}
\bibliography{main}

\begin{thebibliography}{10}\itemsep=-1pt

\bibitem{detr}
Nicolas Carion, Francisco Massa, Gabriel Synnaeve, Nicolas Usunier, Alexander
  Kirillov, and Sergey Zagoruyko.
\newblock End-to-end object detection with transformers.
\newblock In {\em European conference on computer vision}, pages 213--229.
  Springer, 2020.

\bibitem{self}
Mathilde Caron, Hugo Touvron, Ishan Misra, Herv{\'e} J{\'e}gou, Julien Mairal,
  Piotr Bojanowski, and Armand Joulin.
\newblock Emerging properties in self-supervised vision transformers.
\newblock In {\em Proceedings of the IEEE/CVF International Conference on
  Computer Vision}, pages 9650--9660, 2021.

\bibitem{soft3}
Pengguang Chen, Shu Liu, Hengshuang Zhao, and Jiaya Jia.
\newblock Distilling knowledge via knowledge review.
\newblock In {\em Proceedings of the IEEE/CVF Conference on Computer Vision and
  Pattern Recognition}, pages 5008--5017, 2021.

\bibitem{detection7}
Xingyu Chen, Junzhi Yu, Shihan Kong, Zhengxing Wu, and Li Wen.
\newblock Joint anchor-feature refinement for real-time accurate object
  detection in images and videos.
\newblock {\em IEEE Transactions on Circuits and Systems for Video Technology},
  31(2):594--607, 2020.

\bibitem{dai2021dynamic}
Xiyang Dai, Yinpeng Chen, Jianwei Yang, Pengchuan Zhang, Lu Yuan, and Lei
  Zhang.
\newblock Dynamic detr: End-to-end object detection with dynamic attention.
\newblock In {\em Proceedings of the IEEE/CVF International Conference on
  Computer Vision}, pages 2988--2997, 2021.

\bibitem{imagenet}
Jia Deng, Wei Dong, Richard Socher, Li-Jia Li, Kai Li, and Li Fei-Fei.
\newblock Imagenet: A large-scale hierarchical image database.
\newblock In {\em 2009 IEEE conference on computer vision and pattern
  recognition}, pages 248--255. Ieee, 2009.

\bibitem{bert}
Jacob Devlin, Ming-Wei Chang, Kenton Lee, and Kristina Toutanova.
\newblock Bert: Pre-training of deep bidirectional transformers for language
  understanding.
\newblock {\em arXiv preprint arXiv:1810.04805}, 2018.

\bibitem{voc}
Mark Everingham, Luc Van~Gool, Christopher~KI Williams, John Winn, and Andrew
  Zisserman.
\newblock The pascal visual object classes (voc) challenge.
\newblock {\em International journal of computer vision}, 88(2):303--338, 2010.

\bibitem{detection4}
Ross Girshick.
\newblock Fast r-cnn.
\newblock In {\em Proceedings of the IEEE international conference on computer
  vision}, pages 1440--1448, 2015.

\bibitem{vild}
X Gu, T Lin, W Kuo, and Y Cui.
\newblock Zero-shot detection via vision and language knowledge distillation.
  corr abs/2104.13921 (2021).

\bibitem{lvis}
Agrim Gupta, Piotr Dollar, and Ross Girshick.
\newblock Lvis: A dataset for large vocabulary instance segmentation.
\newblock In {\em Proceedings of the IEEE/CVF conference on computer vision and
  pattern recognition}, pages 5356--5364, 2019.

\bibitem{owdetr}
Akshita Gupta, Sanath Narayan, KJ Joseph, Salman Khan, Fahad~Shahbaz Khan, and
  Mubarak Shah.
\newblock Ow-detr: Open-world detection transformer.
\newblock In {\em CVPR}, 2022.

\bibitem{detection3}
Kaiming He, Georgia Gkioxari, Piotr Doll{\'a}r, and Ross Girshick.
\newblock Mask r-cnn.
\newblock In {\em Proceedings of the IEEE international conference on computer
  vision}, pages 2961--2969, 2017.

\bibitem{resnet50}
Kaiming He, Xiangyu Zhang, Shaoqing Ren, and Jian Sun.
\newblock Deep residual learning for image recognition.
\newblock In {\em Proceedings of the IEEE conference on computer vision and
  pattern recognition}, pages 770--778, 2016.

\bibitem{ORE}
K~J Joseph, Salman Khan, Fahad~Shahbaz Khan, and Vineeth~N Balasubramanian.
\newblock Towards open world object detection.
\newblock In {\em 2021 IEEE/CVF Conference on Computer Vision and Pattern
  Recognition (CVPR)}, pages 5826--5836, 2021.

\bibitem{mdetr}
Aishwarya Kamath, Mannat Singh, Yann LeCun, Gabriel Synnaeve, Ishan Misra, and
  Nicolas Carion.
\newblock Mdetr-modulated detection for end-to-end multi-modal understanding.
\newblock In {\em Proceedings of the IEEE/CVF International Conference on
  Computer Vision}, pages 1780--1790, 2021.

\bibitem{soft2}
Guang Li, Ren Togo, Takahiro Ogawa, and Miki Haseyama.
\newblock Soft-label anonymous gastric x-ray image distillation.
\newblock In {\em 2020 IEEE International Conference on Image Processing
  (ICIP)}, pages 305--309. IEEE, 2020.

\bibitem{glip}
Liunian~Harold Li, Pengchuan Zhang, Haotian Zhang, Jianwei Yang, Chunyuan Li,
  Yiwu Zhong, Lijuan Wang, Lu Yuan, Lei Zhang, Jenq-Neng Hwang, et~al.
\newblock Grounded language-image pre-training.
\newblock In {\em Proceedings of the IEEE/CVF Conference on Computer Vision and
  Pattern Recognition}, pages 10965--10975, 2022.

\bibitem{detection6}
Tsung-Yi Lin, Priya Goyal, Ross Girshick, Kaiming He, and Piotr Doll{\'a}r.
\newblock Focal loss for dense object detection.
\newblock In {\em Proceedings of the IEEE international conference on computer
  vision}, pages 2980--2988, 2017.

\bibitem{focal}
Tsung-Yi Lin, Priya Goyal, Ross Girshick, Kaiming He, and Piotr Doll{\'a}r.
\newblock Focal loss for dense object detection.
\newblock In {\em Proceedings of the IEEE international conference on computer
  vision}, pages 2980--2988, 2017.

\bibitem{mscoco}
Tsung-Yi Lin, Michael Maire, Serge Belongie, James Hays, Pietro Perona, Deva
  Ramanan, Piotr Doll{\'a}r, and C~Lawrence Zitnick.
\newblock Microsoft coco: Common objects in context.
\newblock In {\em European conference on computer vision}, pages 740--755.
  Springer, 2014.

\bibitem{swin}
Ze Liu, Yutong Lin, Yue Cao, Han Hu, Yixuan Wei, Zheng Zhang, Stephen Lin, and
  Baining Guo.
\newblock Swin transformer: Hierarchical vision transformer using shifted
  windows.
\newblock In {\em Proceedings of the IEEE/CVF international conference on
  computer vision}, pages 10012--10022, 2021.

\bibitem{detection1}
Yue Lu, Xingyu Chen, Zhengxing Wu, and Junzhi Yu.
\newblock Decoupled metric network for single-stage few-shot object detection.
\newblock {\em IEEE Transactions on Cybernetics}, 2022.

\bibitem{MViTs}
Muhammad Maaz, Hanoona Rasheed, Salman Khan, Fahad~Shahbaz Khan, Rao~Muhammad
  Anwer, and Ming-Hsuan Yang.
\newblock Class-agnostic object detection with multi-modal transformer.
\newblock In {\em Computer Vision--ECCV 2022: 17th European Conference, Tel
  Aviv, Israel, October 23--27, 2022, Proceedings, Part X}, pages 512--531.
  Springer, 2022.

\bibitem{detection2}
Yanwei Pang, Tiancai Wang, Rao~Muhammad Anwer, Fahad~Shahbaz Khan, and Ling
  Shao.
\newblock Efficient featurized image pyramid network for single shot detector.
\newblock In {\em Proceedings of the IEEE/CVF Conference on Computer Vision and
  Pattern Recognition}, pages 7336--7344, 2019.

\bibitem{fasteriod}
Can Peng, Kun Zhao, and Brian~C Lovell.
\newblock Faster ilod: Incremental learning for object detectors based on
  faster rcnn.
\newblock {\em Pattern recognition letters}, 140:109--115, 2020.

\bibitem{clip}
Alec Radford, Jong~Wook Kim, Chris Hallacy, Aditya Ramesh, Gabriel Goh,
  Sandhini Agarwal, Girish Sastry, Amanda Askell, Pamela Mishkin, Jack Clark,
  et~al.
\newblock Learning transferable visual models from natural language
  supervision.
\newblock In {\em International conference on machine learning}, pages
  8748--8763. PMLR, 2021.

\bibitem{detection5}
Joseph Redmon, Santosh Divvala, Ross Girshick, and Ali Farhadi.
\newblock You only look once: Unified, real-time object detection.
\newblock In {\em Proceedings of the IEEE conference on computer vision and
  pattern recognition}, pages 779--788, 2016.

\bibitem{fasterrcnn}
Shaoqing Ren, Kaiming He, Ross Girshick, and Jian Sun.
\newblock Faster r-cnn: Towards real-time object detection with region proposal
  networks.
\newblock {\em Advances in neural information processing systems}, 28, 2015.

\bibitem{GIOU}
Hamid Rezatofighi, Nathan Tsoi, JunYoung Gwak, Amir Sadeghian, Ian Reid, and
  Silvio Savarese.
\newblock Generalized intersection over union: A metric and a loss for bounding
  box regression.
\newblock In {\em Proceedings of the IEEE/CVF conference on computer vision and
  pattern recognition}, pages 658--666, 2019.

\bibitem{iod}
Konstantin Shmelkov, Cordelia Schmid, and Karteek Alahari.
\newblock Incremental learning of object detectors without catastrophic
  forgetting.
\newblock In {\em Proceedings of the IEEE international conference on computer
  vision}, pages 3400--3409, 2017.

\bibitem{soft1}
Ilia Sucholutsky and Matthias Schonlau.
\newblock Soft-label dataset distillation and text dataset distillation.
\newblock In {\em 2021 International Joint Conference on Neural Networks
  (IJCNN)}, pages 1--8. IEEE, 2021.

\bibitem{two-branch}
Yan Wu, Xiaowei Zhao, Yuqing Ma, Duorui Wang, and Xianglong Liu.
\newblock Two-branch objectness-centric open world detection.
\newblock In {\em Proceedings of the 3rd International Workshop on
  Human-Centric Multimedia Analysis}, pages 35--40, 2022.

\bibitem{UC-OWOD}
Zhiheng Wu, Yue Lu, Xingyu Chen, Zhengxing Wu, Liwen Kang, and Junzhi Yu.
\newblock Uc-owod: Unknown-classified open world object detection.
\newblock In {\em Computer Vision--ECCV 2022: 17th European Conference, Tel
  Aviv, Israel, October 23--27, 2022, Proceedings, Part X}, pages 193--210.
  Springer, 2022.

\bibitem{detclip}
Lewei Yao, Jianhua Han, Youpeng Wen, Xiaodan Liang, Dan Xu, Wei Zhang, Zhenguo
  Li, Chunjing Xu, and Hang Xu.
\newblock Detclip: Dictionary-enriched visual-concept paralleled pre-training
  for open-world detection.
\newblock {\em arXiv preprint arXiv:2209.09407}, 2022.

\bibitem{OCPL}
Jinan Yu, Liyan Ma, Zhenglin Li, Yan Peng, and Shaorong Xie.
\newblock Open-world object detection via discriminative class prototype
  learning.
\newblock In {\em 2022 IEEE International Conference on Image Processing
  (ICIP)}, pages 626--630. IEEE, 2022.

\bibitem{glipv2}
Haotian Zhang, Pengchuan Zhang, Xiaowei Hu, Yen-Chun Chen, Liunian~Harold Li,
  Xiyang Dai, Lijuan Wang, Lu Yuan, Jenq-Neng Hwang, and Jianfeng Gao.
\newblock Glipv2: Unifying localization and vision-language understanding.
\newblock In {\em Advances in Neural Information Processing Systems}, 2022.

\bibitem{Revisiting}
Xiaowei Zhao, Xianglong Liu, Yifan Shen, Yuqing Ma, Yixuan Qiao, and Duorui
  Wang.
\newblock Revisiting open world object detection.
\newblock {\em arXiv preprint arXiv:2201.00471}, 2022.

\bibitem{ddetr}
Xizhou Zhu, Weijie Su, Lewei Lu, Bin Li, Xiaogang Wang, and Jifeng Dai.
\newblock Deformable detr: Deformable transformers for end-to-end object
  detection.
\newblock {\em arXiv preprint arXiv:2010.04159}, 2020.

\bibitem{zou2019object}
Zhengxia Zou, Zhenwei Shi, Yuhong Guo, and Jieping Ye.
\newblock Object detection in 20 years: A survey.
\newblock {\em arXiv preprint arXiv:1905.05055}, 2019.

\end{thebibliography}
}
%---------------------------------------------------------------------------------------
\appendix
\clearpage
\onecolumn{%
\renewcommand\onecolumn[1][]{#1}%
\maketitle
\section{Additional Experimental Results}
\subsection{Evaluation on WI and A-OSE metrics}
Table \ref{table6} shows a comparison of the different open-world object detection(OWOD) methods on the MS-COCO split in terms of unknown recall (U-recall), wilderness impact (WI), and absolute open-set error (A-OSE). U-Recall measures the models’ ability to detect unknown objects and indicates the degree of unknown objects that are detected by an OWOD method. Meanwhile, the WI and A-OSE measure the model’s confusion in predicting an unknown instance as a known class. Specifically, WI measures the effect the unknown detections have on the model’s precision. The calculation formula is as follows: 
\begin{equation}
\mathrm{WI}=\frac{P_{\mathcal{K}}}{P_{\mathcal{K} \cup \mathcal{U}}}-1
\end{equation}
Where ${P}_\mathcal{K}$ is the prediction on known classes and ${P}_{\mathcal{K}\cup  \mathcal{U}}$ is the prediction on known and unknown classes. A-OSE devotes the total number of unknown instances detected as known classes. Both WI and A-OSE indicate the degree of confusion in predicting the known classes in the presence of unknown instances. Furthermore, U-Recall directly measures the model’s ability to retrieve the unknown instances.
\begin{table*}[htbp]
\renewcommand\arraystretch{1.25}
\centering
\resizebox{0.8\textwidth}{!}{
\begin{tabular}{l| ccc| ccc |ccc }\toprule

\multicolumn{1}{c}{Task IDs} & \multicolumn{3}{|c}{Task 1}                                                        & \multicolumn{3}{|c}{Task 2}                                                        & \multicolumn{3}{|c}{Task 3}                                                        \\ \midrule
\multicolumn{1}{c}{Metrics $\rightarrow$} & \multicolumn{1}{|c}{ \cellcolor[HTML]{fcfce0}U-Recall$(\uparrow)$} & \multicolumn{1}{c}{\cellcolor[HTML]{ffedec}WI$(\downarrow)$} & \multicolumn{1}{c}{\cellcolor[HTML]{ccfecb}A-OSE$(\downarrow)$} & \multicolumn{1}{|c}{ \cellcolor[HTML]{fcfce0}U-Recall$(\uparrow)$} & \multicolumn{1}{c}{\cellcolor[HTML]{ffedec}WI$(\downarrow)$} & \multicolumn{1}{c}{\cellcolor[HTML]{ccfecb}A-OSE$(\downarrow)$} & \multicolumn{1}{|c}{ \cellcolor[HTML]{fcfce0}U-Recall$(\uparrow)$} & \multicolumn{1}{c}{\cellcolor[HTML]{ffedec}WI$(\downarrow)$} & \multicolumn{1}{c}{\cellcolor[HTML]{ccfecb}A-OSE$(\downarrow)$} \\ \midrule
% ORE-EBUI                     &                              &                        &                           &                              &                        &                           &                              &                        &                           \\
OW-DETR                      & 5.7                          & 0.0458                 & 19815                     & 6.2                          & 0.0499                 & 19749                     & 6.9                          & 0.0248                 & 9233                      \\
Ours                         & 60.9                         & 0.0449                 & 5028                      & 60.0                         & 0.0344                 & 5359                      & 58.6                         & 0.0214                 & 3233                      \\

\bottomrule 
\end{tabular}}
\setlength{\abovecaptionskip}{0.15cm}
\caption{Uknown Object Confusion on MS-COCO split.The comparison is shown in terms of wilderness impact (WI), absolute open set error (A-OSE) and unknown class recall (U-Recall). The unknown recall (U-Recall) metric quantifies a model’s ability to retrieve the unknown object instances. Ours  chieves improved WI, A-OSE and U-Recall over OW-DETR across tasks, thereby indicating less confusion in detecting unknown instances as known classes with higher unknown instance detection capabilities. } 
\label{table6}
\end{table*}

\begin{table*}[htbp]
\renewcommand\arraystretch{1.25}
% \centering

\resizebox{\textwidth}{!}{
\begin{tabular}{lccccccccccccccccccccc}
\toprule
10 + 10 setting  & aero & cycle & bird & boat & bottle & bus  & car  & cat  & chair & cow  & table  & dog  & horse  & bike   & person & plant  & sheep  & sofa   & train  & tv & mAP  \\
\midrule
ILOD & 69.9 & 70.4  & 69.4 & 54.3 & 48   & 68.7 & 78.9 & 68.4 & 45.5  & 58.1 & \cellcolor[HTML]{FAF4E2}59.7&\cellcolor[HTML]{FAF4E2}72.7 &\cellcolor[HTML]{FAF4E2}73.5 &\cellcolor[HTML]{FAF4E2}73.2 &\cellcolor[HTML]{FAF4E2}66.3 &\cellcolor[HTML]{FAF4E2}29.5 &\cellcolor[HTML]{FAF4E2}63.4 &\cellcolor[HTML]{FAF4E2}61.6 &\cellcolor[HTML]{FAF4E2}69.3 &\cellcolor[HTML]{FAF4E2}62.2 & 63.2 \\
Faster ILOD  & 72.8 & 75.7  &  71.2  &  60.5  &  61.7  &  70.4  &  83.3  &  76.6  &  53.1  &  72.3  &\cellcolor[HTML]{FAF4E2} 36.7  &\cellcolor[HTML]{FAF4E2} 70.9  &\cellcolor[HTML]{FAF4E2} 66.8  &\cellcolor[HTML]{FAF4E2} 67.6  &\cellcolor[HTML]{FAF4E2} 66.1  &\cellcolor[HTML]{FAF4E2}24.7  &\cellcolor[HTML]{FAF4E2} 63.1  &\cellcolor[HTML]{FAF4E2} 48.1  &\cellcolor[HTML]{FAF4E2} 57.1  &\cellcolor[HTML]{FAF4E2} 43.6  & 62.1 \\
ORE - (CC + EBUI)  &53.3& 69.2& 62.4& 51.8& 52.9& 73.6& 83.7& 71.7& 42.8& 66.8& \cellcolor[HTML]{FAF4E2}46.8&\cellcolor[HTML]{FAF4E2}59.9&\cellcolor[HTML]{FAF4E2}65.5&\cellcolor[HTML]{FAF4E2}66.1&\cellcolor[HTML]{FAF4E2}68.6&\cellcolor[HTML]{FAF4E2}29.8&\cellcolor[HTML]{FAF4E2}55.1&\cellcolor[HTML]{FAF4E2}51.6&\cellcolor[HTML]{FAF4E2}65.3&\cellcolor[HTML]{FAF4E2}51.5& 59.4  \\
ORE - EBUI &63.5& 70.9& 58.9& 42.9& 34.1& 76.2& 80.7& 76.3& 34.1& 66.1&\cellcolor[HTML]{FAF4E2}56.1&\cellcolor[HTML]{FAF4E2}70.4&\cellcolor[HTML]{FAF4E2}80.2&\cellcolor[HTML]{FAF4E2}72.3&\cellcolor[HTML]{FAF4E2}81.8&\cellcolor[HTML]{FAF4E2}42.7&\cellcolor[HTML]{FAF4E2}71.6&\cellcolor[HTML]{FAF4E2}68.1&\cellcolor[HTML]{FAF4E2}77&\cellcolor[HTML]{FAF4E2}67.7& 64.5  \\

OW - DETR &75.4&63.9&57.9&50.0&52.0&70.9&79.5&72.4&44.3&57.9&\cellcolor[HTML]{FAF4E2}59.7&\cellcolor[HTML]{FAF4E2}73.5&\cellcolor[HTML]{FAF4E2}77.7&\cellcolor[HTML]{FAF4E2}75.2&\cellcolor[HTML]{FAF4E2}76.2&\cellcolor[HTML]{FAF4E2}44.9&\cellcolor[HTML]{FAF4E2}68.8&\cellcolor[HTML]{FAF4E2}65.4&\cellcolor[HTML]{FAF4E2}79.3&\cellcolor[HTML]{FAF4E2}69.0   &65.7  \\\midrule
\textbf{Ours} & 77.1& 72.3&74.5&53.4&57.4& 78.1& 78.7 &83.9& 46.2 & 71.4  &\cellcolor[HTML]{FAF4E2}59.5  &\cellcolor[HTML]{FAF4E2}77.4   &\cellcolor[HTML]{FAF4E2}73.3   &\cellcolor[HTML]{FAF4E2}76.6   &\cellcolor[HTML]{FAF4E2}73.3   &\cellcolor[HTML]{FAF4E2}39.7   &\cellcolor[HTML]{FAF4E2}70.6   &\cellcolor[HTML]{FAF4E2}59.0 &\cellcolor[HTML]{FAF4E2}78.4   &\cellcolor[HTML]{FAF4E2}70.9   &68.6 \\\midrule\midrule

15 + 5 setting& aero & cycle & bird & boat & bottle & bus  & car  & cat  & chair & cow  & table  & dog  & horse  & bike & person & plant  & sheep  & sofa & train  & tv & mAP  \\\midrule
ILOD & 70.5 & 79.2 &68.8& 59.1& 53.2& 75.4& 79.4& 78.8& 46.6& 59.4& 59 &75.8& 71.8& 78.6& 69.6&\cellcolor[HTML]{FAF4E2}33.7&\cellcolor[HTML]{FAF4E2}61.5&\cellcolor[HTML]{FAF4E2}63.1 &\cellcolor[HTML]{FAF4E2}71.7&\cellcolor[HTML]{FAF4E2}62.2& 65.8 \\
Faster ILOD  &66.5& 78.1& 71.8& 54.6& 61.4& 68.4& 82.6& 82.7& 52.1& 74.3& 63.1& 78.6& 80.5& 78.4& 80.4&\cellcolor[HTML]{FAF4E2}36.7&\cellcolor[HTML]{FAF4E2}61.7&\cellcolor[HTML]{FAF4E2}59.3&\cellcolor[HTML]{FAF4E2}67.9&\cellcolor[HTML]{FAF4E2}59.1 &67.9 \\
ORE - (CC + EBUI)  &65.1& 74.6& 57.9& 39.5 &36.7& 75.1& 80 &73.3& 37.1& 69.8& 48.8& 69 &77.5& 72.8& 76.5&\cellcolor[HTML]{FAF4E2}34.4&\cellcolor[HTML]{FAF4E2}62.6&\cellcolor[HTML]{FAF4E2}56.5&\cellcolor[HTML]{FAF4E2}80.3&\cellcolor[HTML]{FAF4E2}65.7& 62.6 \\
ORE - EBUI &75.4& 81& 67.1& 51.9& 55.7& 77.2& 85.6& 81.7& 46.1& 76.2& 55.4& 76.7& 86.2& 78.5& 82.1&\cellcolor[HTML]{FAF4E2}32.8&\cellcolor[HTML]{FAF4E2}63.6&\cellcolor[HTML]{FAF4E2}54.7&\cellcolor[HTML]{FAF4E2}77.7&\cellcolor[HTML]{FAF4E2}64.6 &68.5 \\
OW - DETR &78.0&80.7&79.4&70.4&58.8&65.1&84.0&86.2&56.5&76.7&62.4&84.8&85.0&81.8&81.0&\cellcolor[HTML]{FAF4E2}34.3&\cellcolor[HTML]{FAF4E2}48.2&\cellcolor[HTML]{FAF4E2}57.9&\cellcolor[HTML]{FAF4E2}62.0&\cellcolor[HTML]{FAF4E2}57.0 &69.4\\\midrule
\textbf{Ours} &79.5&85.1&83.1 &73.1 &62.5 &68.7 &83.0 &88.4 &55.5 &78.3&69.7 &83.0 &86.6 &73.2&78.8 &\cellcolor[HTML]{FAF4E2}30.8&\cellcolor[HTML]{FAF4E2}67.6&\cellcolor[HTML]{FAF4E2}60.8&\cellcolor[HTML]{FAF4E2}76.0&\cellcolor[HTML]{FAF4E2}58.7 &72.1 \\\midrule\midrule

19 + 1 setting  & aero & cycle & bird & boat & bottle & bus  & car  & cat  & chair & cow  & table  & dog  & horse  & bike   & person & plant  & sheep  & sofa   & train  & tv & mAP  \\\midrule
ILOD & 69.4&79.3&69.5&57.4&45.4&78.4&79.1&80.5&45.7&76.3&64.8&77.2&80.8&77.5&70.1&42.3&67.5&64.4&76.7&\cellcolor[HTML]{FAF4E2}62.7&68.2  \\
Faster ILOD   &64.2&74.7&73.2&55.5&53.7&70.8&82.9&82.6&51.6&79.7&58.7&78.8&81.8&75.3&77.4&43.1&73.8&61.7&69.8&\cellcolor[HTML]{FAF4E2}61.1&68.5  \\
ORE - (CC + EBUI)  &60.7&78.6&61.8&45&43.2&75.1&82.5&75.5&42.4&75.1&56.7&72.9&80.8&75.4&77.7&37.8&72.3&64.5&70.7&\cellcolor[HTML]{FAF4E2}49.9&64.9 \\
ORE - EBUI   &67.3&76.8&60&48.4&58.8&81.1&86.5&75.8&41.5&79.6&54.6&72.8&85.9&81.7&82.4&44.8&75.8&68.2&75.7&\cellcolor[HTML]{FAF4E2}60.1&68.8\\

OW - DETR &82.2&80.7&73.9&56.0&58.6&72.1&82.4&79.6&48.0&72.8&64.2&83.3&83.1&82.3&78.6&42.1&65.5&55.4&82.9&\cellcolor[HTML]{FAF4E2}60.1  &70.2  \\\midrule

\textbf{Ours} &83.6 &85.7 &77.1 &61.5&58.9 &74.3
&86.3 &81.5 &52.2 &78.4 &71.4 &81.9 &84.6&80.2 &80.8 &39.9 &68.3&63.3 &84.6&\cellcolor[HTML]{FAF4E2}63.0  &72.9  \\ \bottomrule
\end{tabular}
}
\caption{The detailed comparison with existing approaches on PASCAL VOC. We only use our open-world object detector without assistance.  Evaluation is performed on three standard settings, where a group of classes (10, 5 and last class) are introduced incrementally to a detector trained on the remaining classes (10,15 and 19). Our model performs favourably against existing approaches on all three settings.} 
\label{iod}
\end{table*}
\subsection{Additional Incremental Object Detection}
Table.\ref{iod} shows a detailed comparison of CAT with existing approaches on PASCAL VOC. Evaluation is performed on three standard settings, where a group of classes (10, 5 and last class) are introduced incrementally to a detector trained on the remaining classes (10,15 and 19). Our model performs favorably against existing approaches on all three settings, illustrating the power of localization identification cascade detection transformer for incremental objection detection.
\section{Additional Quantitative Results}
We provide more visualization results to compare our model with SOTA model, the large pre-trained grounded language-images model and our model with pseudo labels. Fig.\ref{SOTA1},\ref{SOTA2},\ref{SOTA3} and \ref{SOTA4} illustrate the comparison between SOTA model and our model on Task 1,2,3,4, respectively. In Fig.\ref{GLIPT1},\ref{GLIPT2},\ref{GLIPT3} and \ref{GLIPT4}, we illustrate the comparison between the large pre-trained grounded language-image model and our model on Task 1,2,3,4, respectively. We ablate the pseudo labels in Fig.\ref{NOT1},\ref{NOT2},\ref{NOT3} and \ref{NOT4}. Fig.\ref{IOD} shows the ability of our model to incrementally learn new objects.

\begin{figure*}[htbp]
  \centering
  \includegraphics[width = 0.7\textwidth]{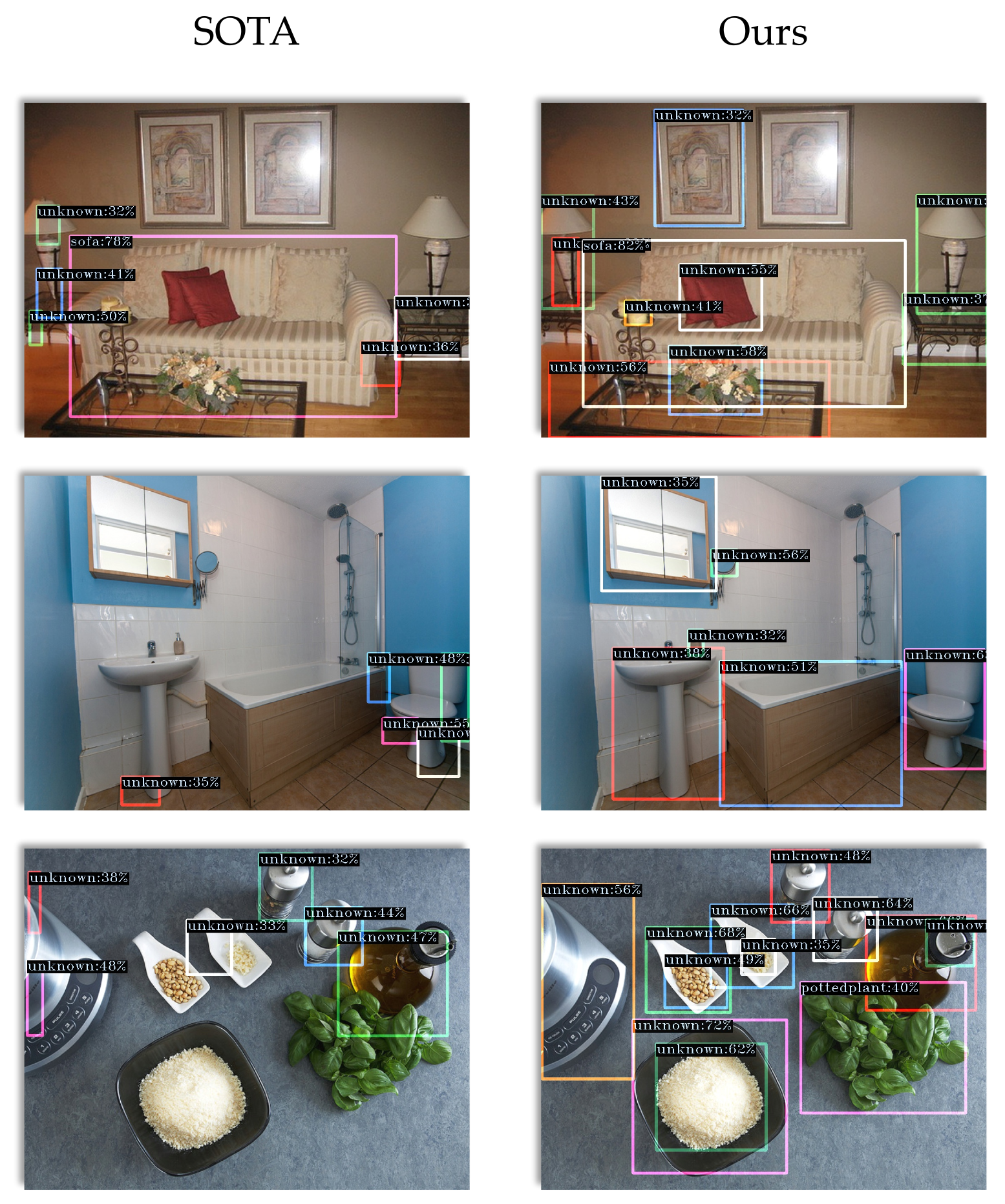}
  \caption{Visualization comparison between the SOTA model and our model on Task 1.}
  \label{SOTA1}
\end{figure*}

\begin{figure*}[htbp]
  \centering
  \includegraphics[width = 0.8\textwidth]{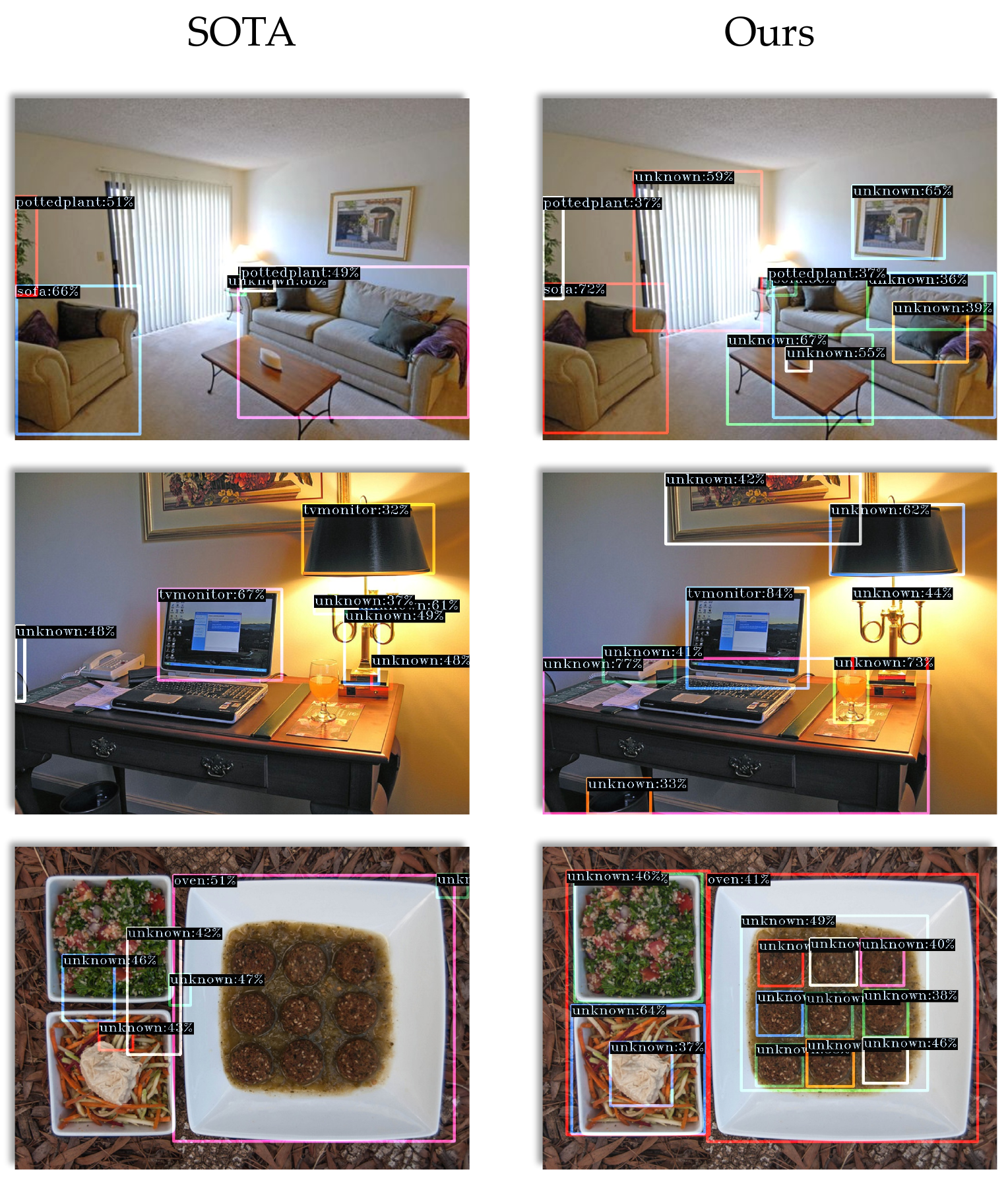}
  \caption{Visualization comparison between the SOTA model and our model on Task 2.}
  \label{SOTA2}
\end{figure*}

\begin{figure*}[htbp]
  \centering
  \includegraphics[width = 0.8\textwidth]{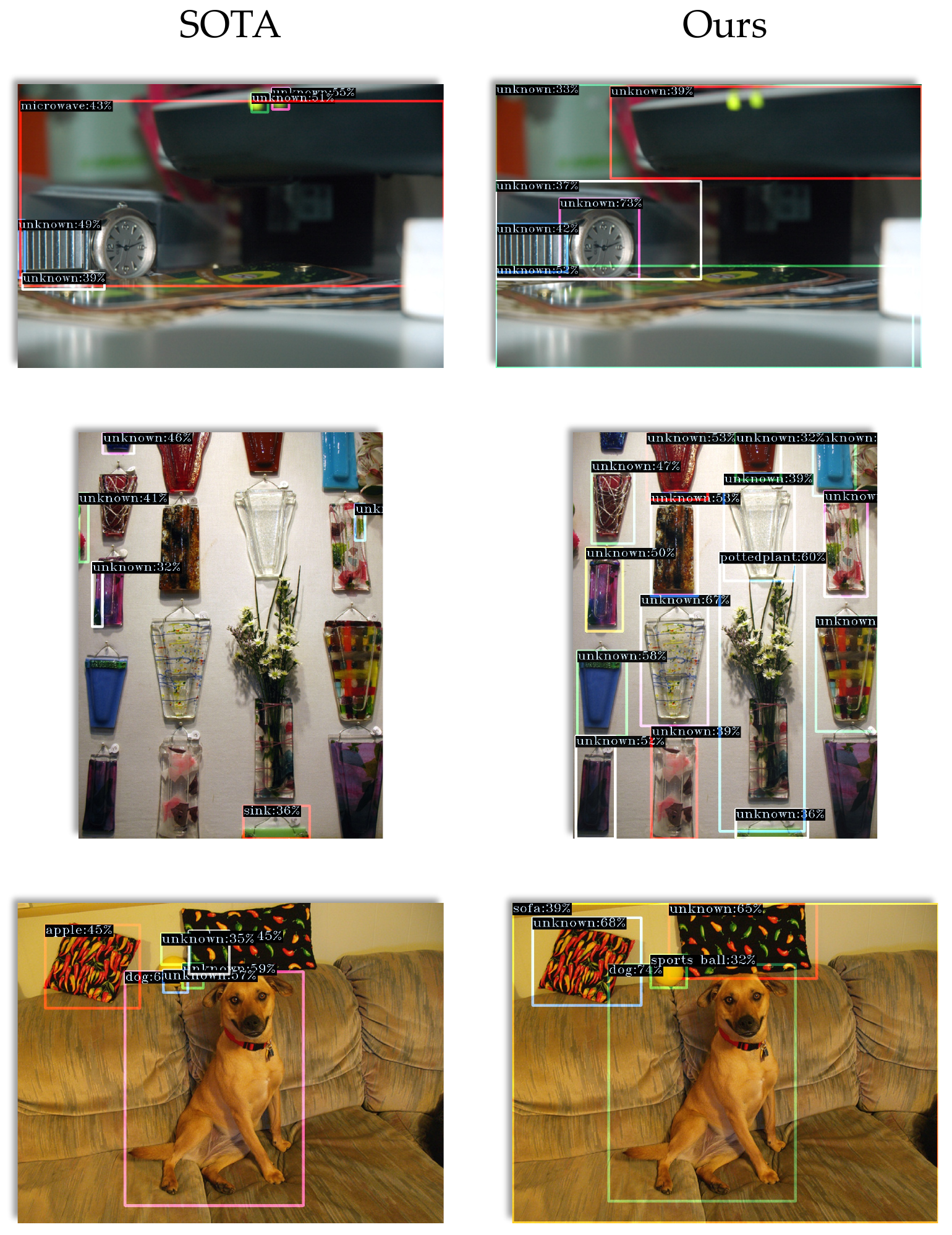}
  \caption{Visualization comparison between the SOTA model and our model on Task 3.}
  \label{SOTA3}
\end{figure*}

\begin{figure*}[htbp]
  \centering
  \includegraphics[width = 0.8\textwidth]{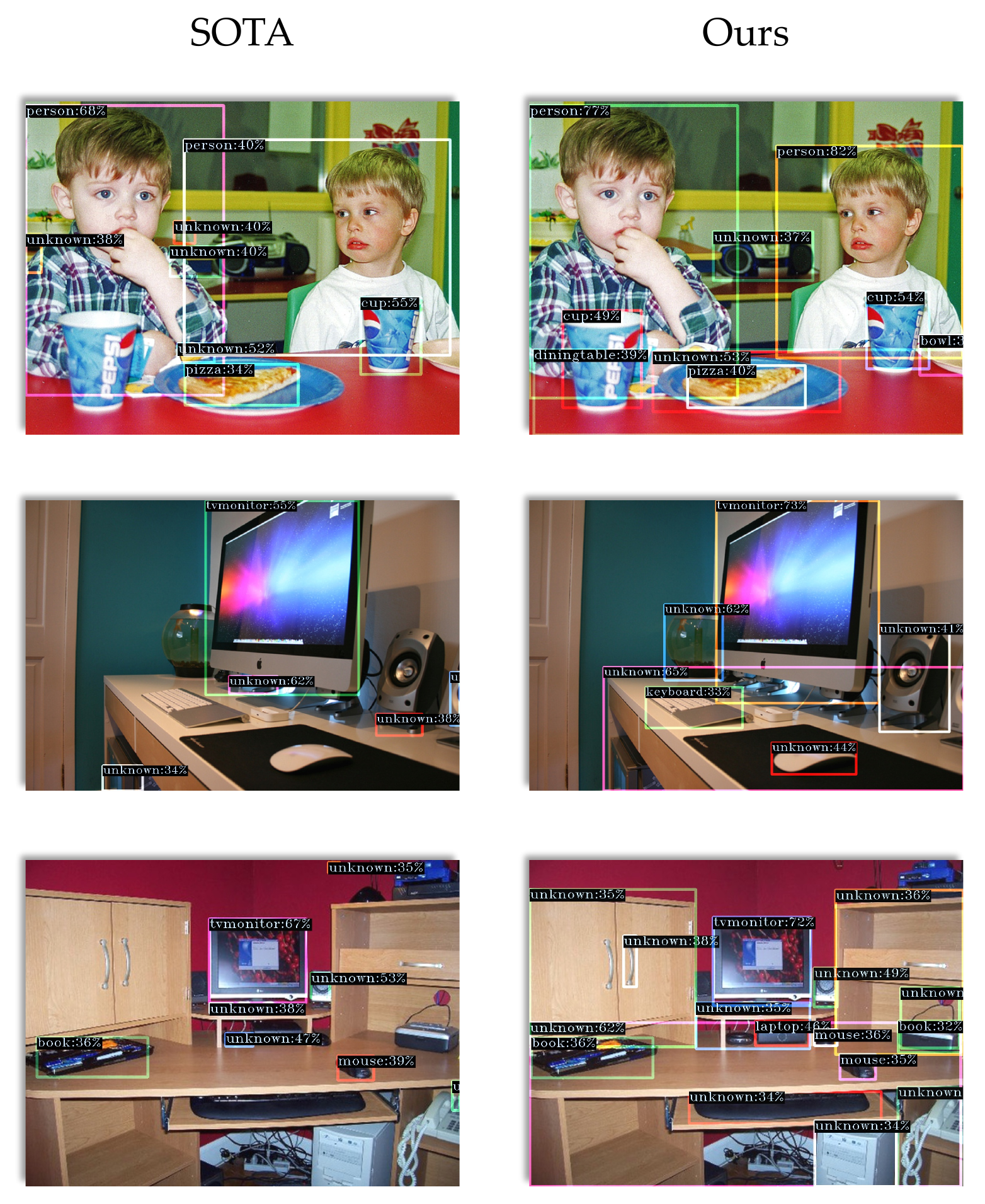}
  \caption{Visualization comparison between the SOTA model and our model on Task 4.}
  \label{SOTA4}
\end{figure*}

\begin{figure*}[htbp]
  \centering
  \includegraphics[width = 0.8\textwidth]{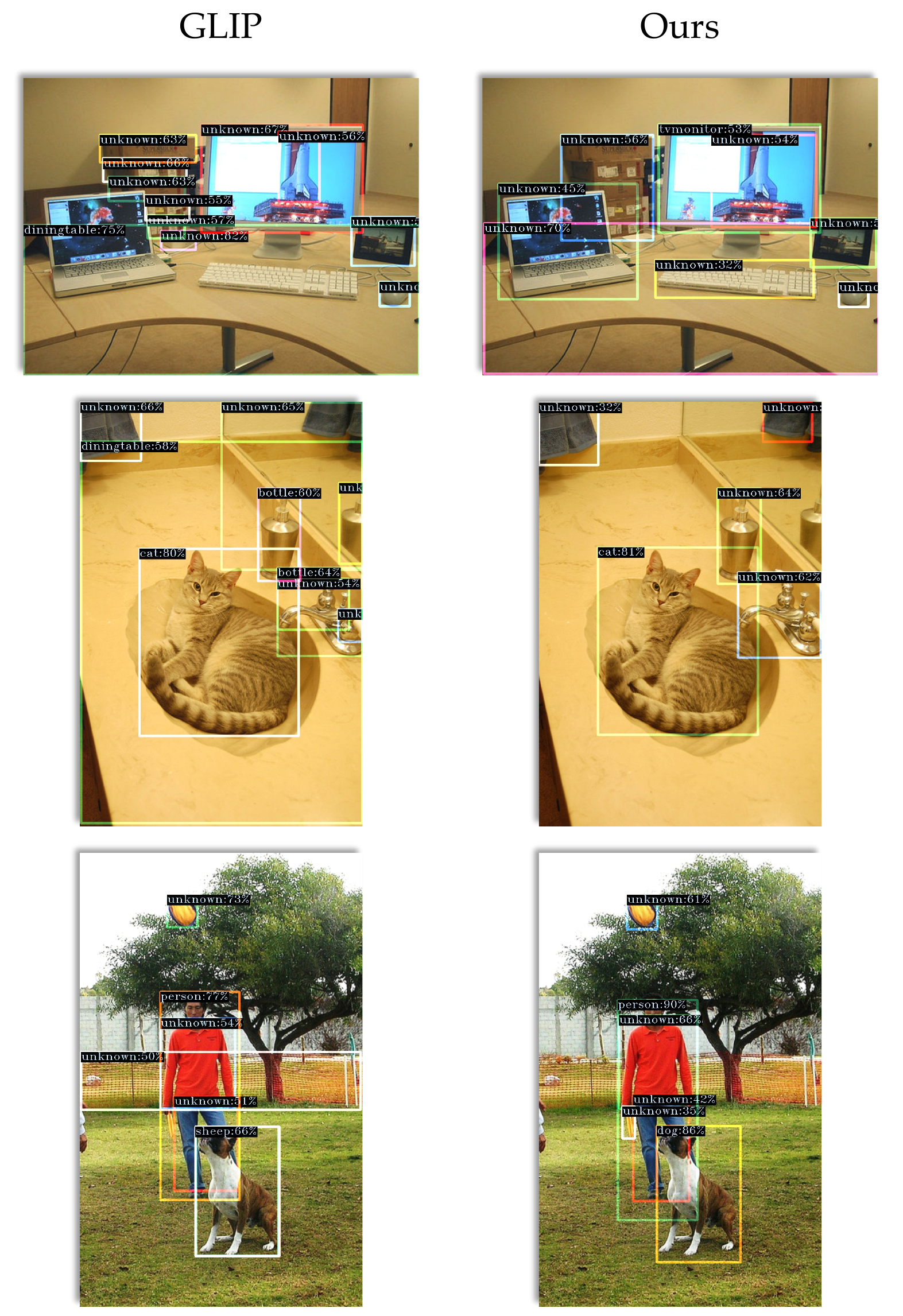}
  \caption{Visualization comparison between the large grounded language-image model and our model on Task 1.}
  \label{GLIPT1}
\end{figure*}

\begin{figure*}[htbp]
  \centering
  \includegraphics[width = 0.8\textwidth]{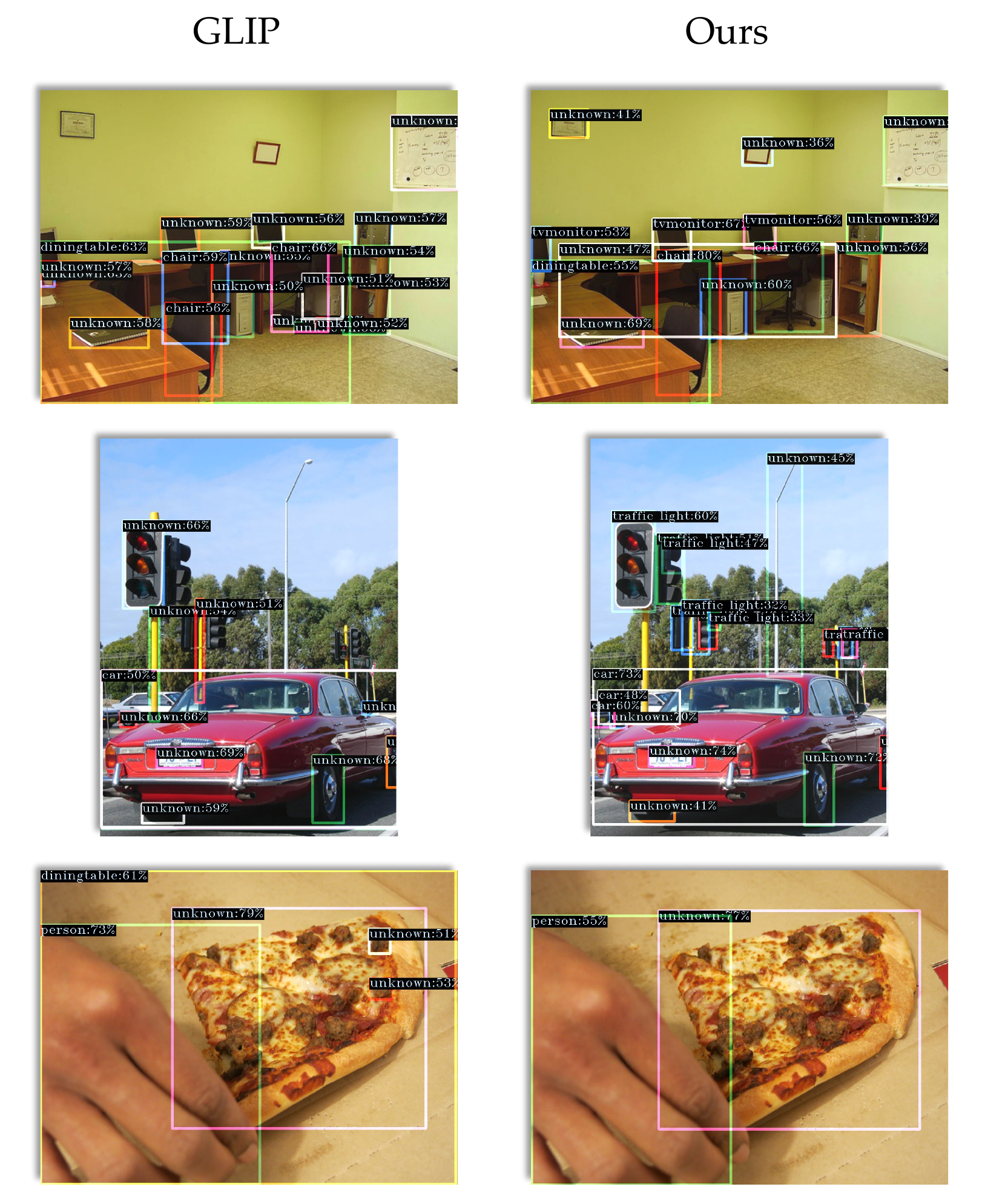}
  \caption{Visualization comparison between the large grounded language-image model and our model on Task 2.}
  \label{GLIPT2}
\end{figure*}

\begin{figure*}[htbp]
  \centering
  \includegraphics[width = 0.8\textwidth]{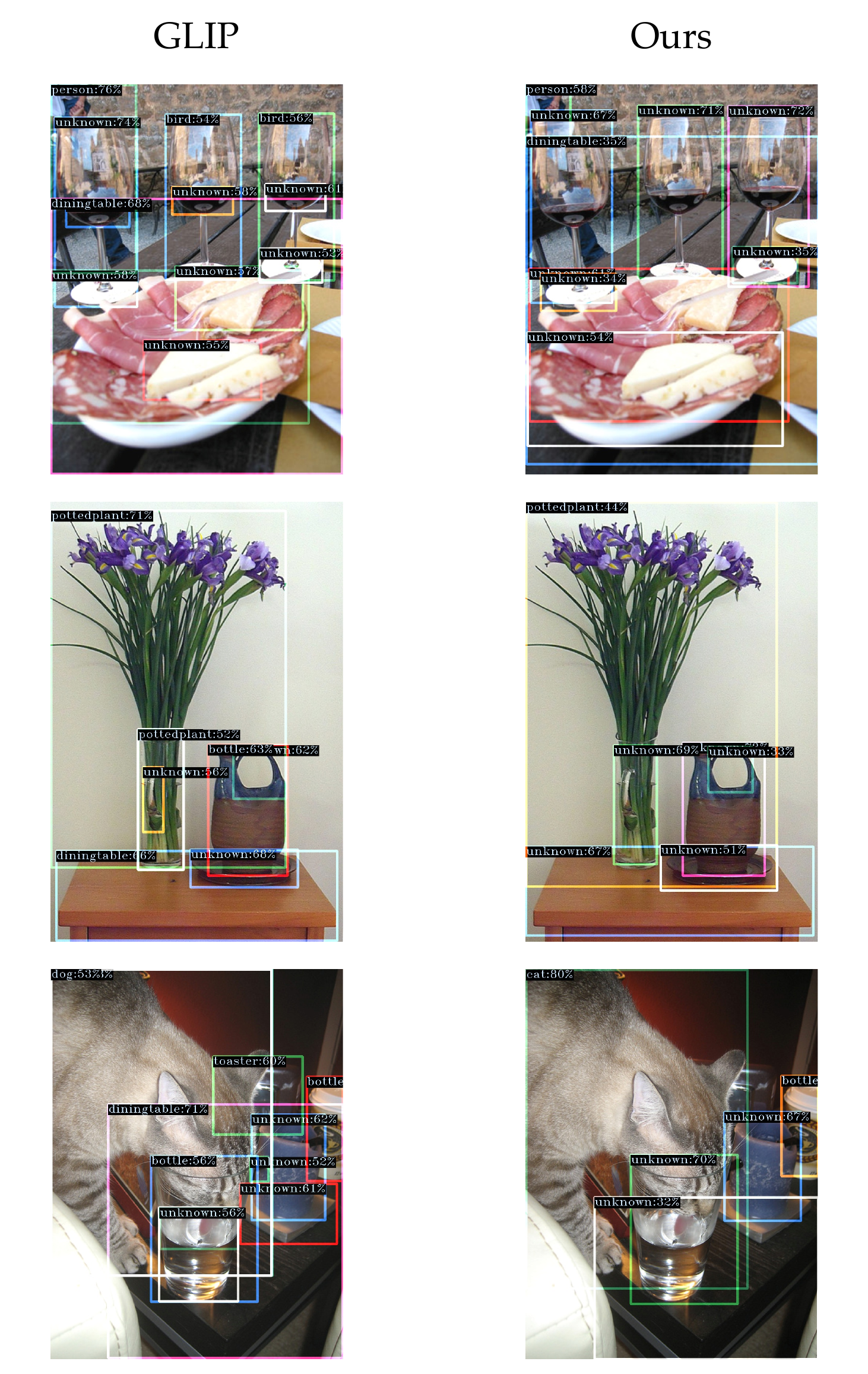}
  \caption{Visualization comparison between the large grounded language-image model and our model on Task 3.}
  \label{GLIPT3}
\end{figure*}

\begin{figure*}[htbp]
  \centering
  \includegraphics[width = 0.8\textwidth]{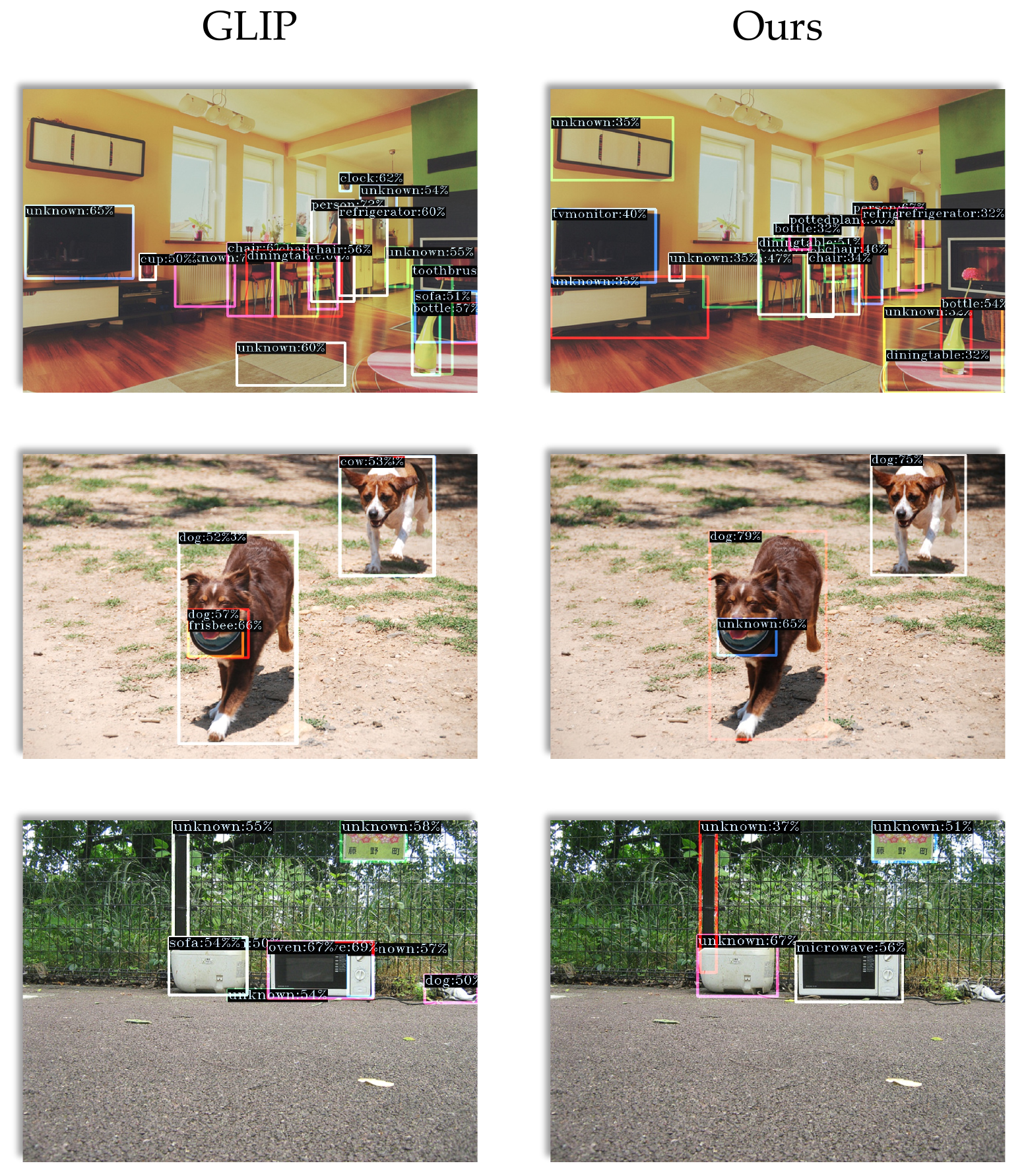}
  \caption{Visualization comparison between the large grounded language-image model and our model on Task 4.}
  \label{GLIPT4}
\end{figure*}

\begin{figure*}[htbp]
  \centering
  \includegraphics[width = 0.8\textwidth]{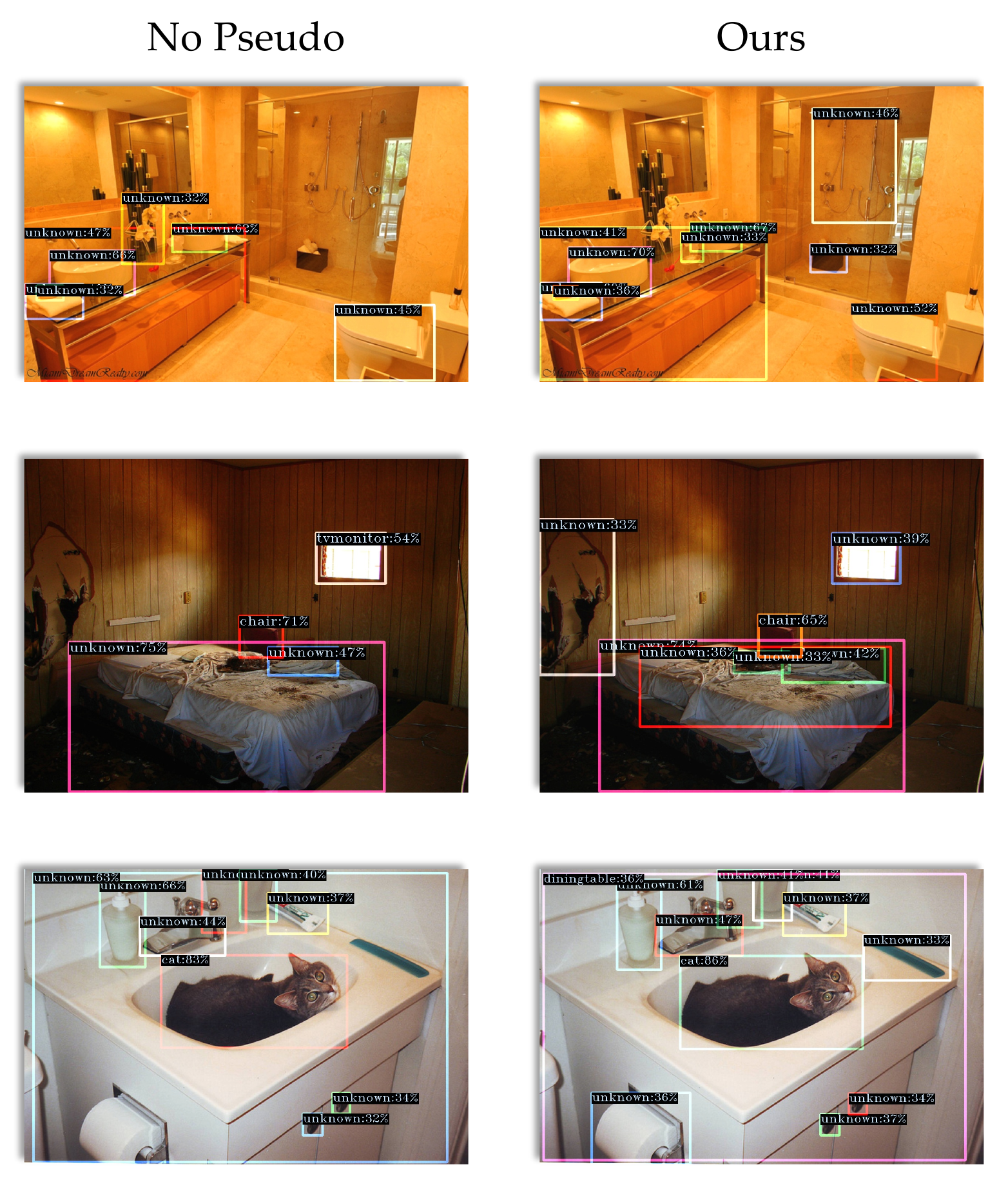}
  \caption{Visualization comparison between our model trained without pseudo labels and our model on Task 1.}
  \label{NOT1}
\end{figure*}

\begin{figure*}[htbp]
  \centering
  \includegraphics[width = 0.8\textwidth]{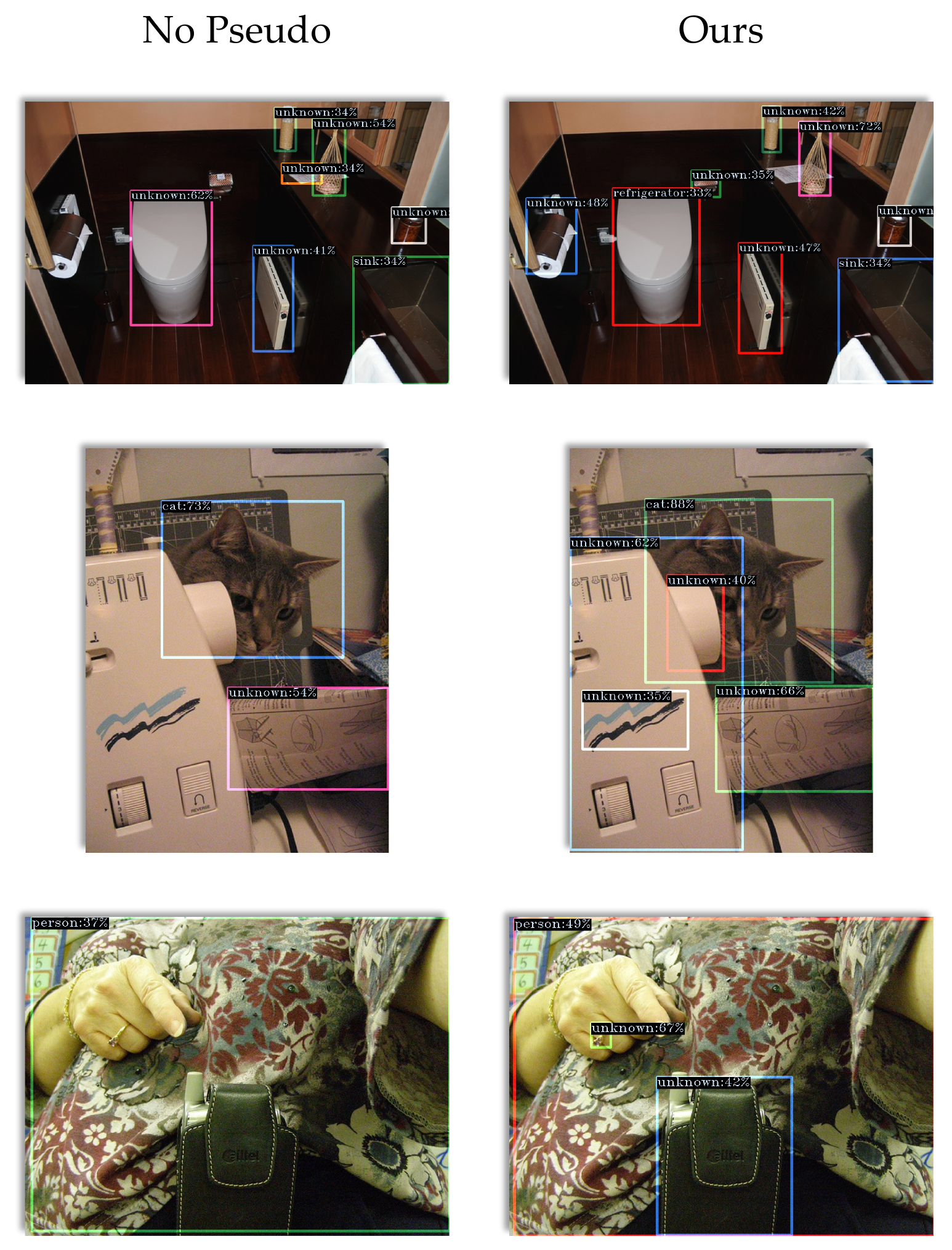}
  \caption{Visualization comparison between our model trained without pseudo labels and our model on Task 2.}
  \label{NOT2}
\end{figure*}

\begin{figure*}[htbp]
  \centering
  \includegraphics[width = 0.8\textwidth]{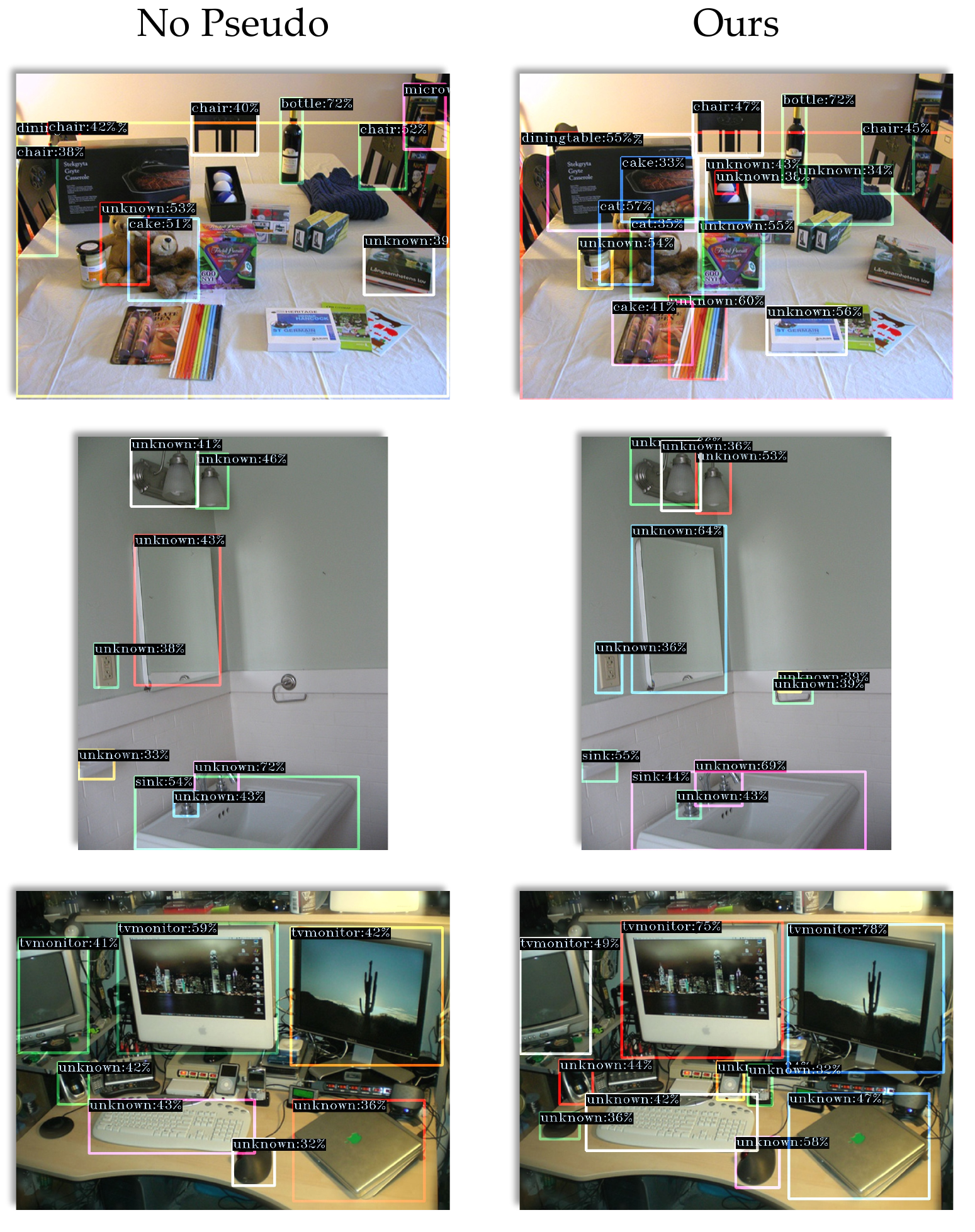}
  \caption{Visualization comparison between our model trained without pseudo labels and our model on Task 3.}
  \label{NOT3}
\end{figure*}

\begin{figure*}[htbp]
  \centering
  \includegraphics[width = 0.8\textwidth]{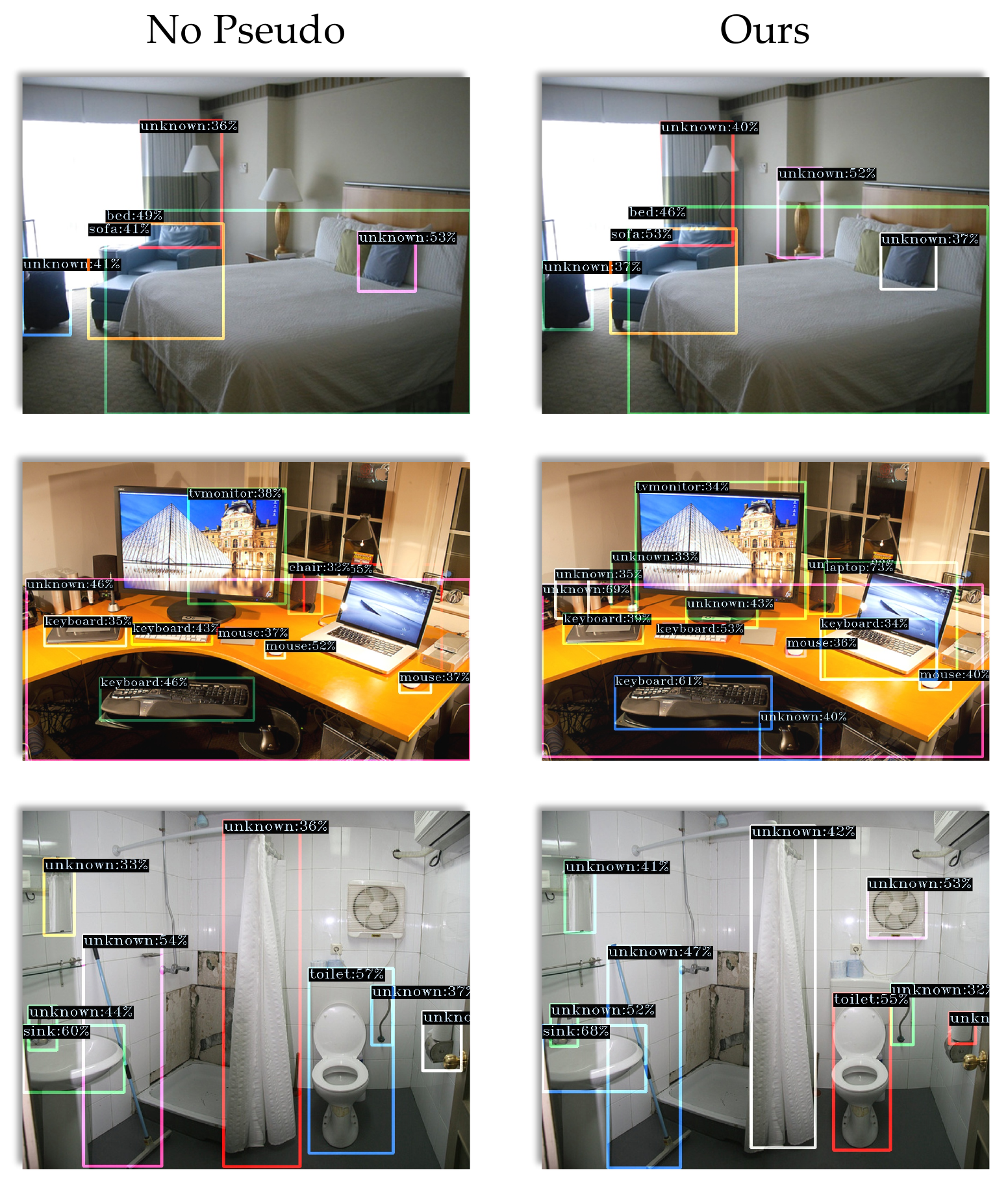}
  \caption{Visualization comparison between our model trained without pseudo labels and our model on Task 4.}
  \label{NOT4}
\end{figure*}

\begin{figure*}[htbp]
  \centering
  \includegraphics[width = 0.8\textwidth]{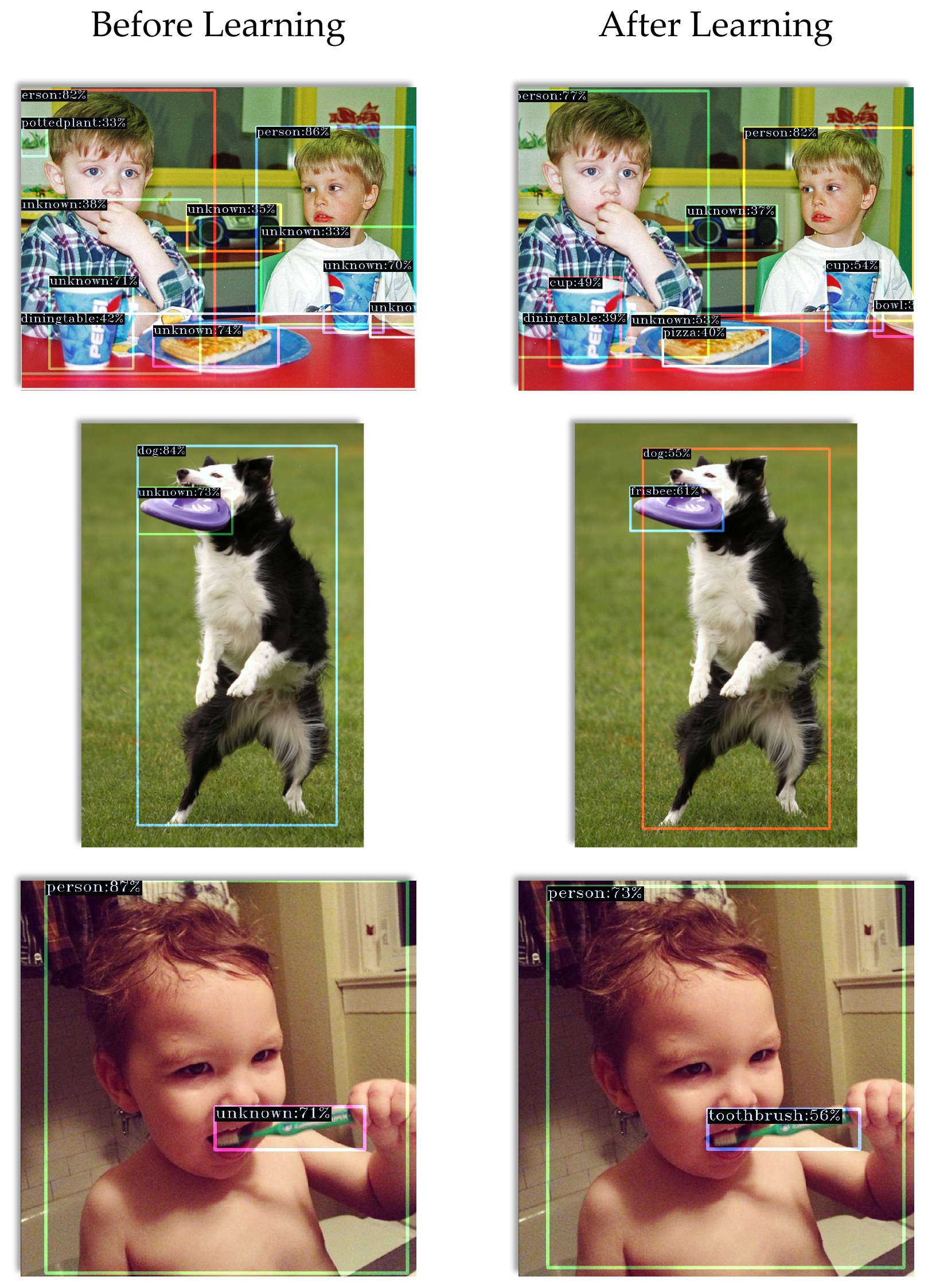}
  \caption{Visualization result for the incremental object detection.}
  \label{IOD}
\end{figure*}

}

\end{document}